\documentclass[10pt,notitlepage]{article}
\usepackage{epsfig}
\usepackage{amssymb}

\newcommand{\be}{\begin{equation}}
\newcommand{\ee}{\end{equation}}
\newcommand{\bea}{\begin{eqnarray}}
\newcommand{\eea}{\end{eqnarray}}

\usepackage{bm}
\usepackage{color}
\usepackage{amsmath}
\usepackage{graphicx}
\usepackage{amssymb}
\usepackage{cite}

\begin{document}


\title{A fast noise filtering algorithm for time series prediction using
recurrent neural networks}
\author{Boris  Rubinstein,
\\Stowers Institute for Medical Research
\\1000 50$^{}\mbox{th}$ St., Kansas City, MO 64110, U.S.A.}
\date{\today}
\maketitle

\begin{abstract}
Recent researches demonstrate that
prediction of time series by recurrent neural networks (RNNs) based on the 
noisy input generates a {\it smooth} anticipated trajectory. We examine 
the internal dynamics of RNNs and establish a set of conditions required for such behavior.
Based on this analysis we propose a new approximate algorithm and show that it
significantly speeds up the predictive process without loss of accuracy and
demonstrates increased robustness in neuroscience context.
\end{abstract}


\section{Introduction}

Recurrent neural networks (RNNs)  due to
their ability to process sequences of data have found applications in many fields 
of science, engineering and humanities, including speech, handwriting and 
human action recognition, automatic translation, robot control etc.
One of the RNN application is time series prediction used
in analysis of  business and financial data, anomaly detection, weather forecast.
A large number of different architectures were discussed recently and 
the flow of new modifications of standard RNN continues to increase and
all these architectures share some common features inherited from the 
basic systems.

Trajectory prediction based on incomplete or noisy data is one of the 
most amazing features of organism brains that allows living creatures to
survive in complex and mostly unfriendly environment. 
A large number of mathematical algorithms developed for this purpose
have many applications in multiple engineering field, e.g.,
development of guidance systems, self-driving vehicles, motor control etc.
\cite{TrajPredReview2018}.

It was shown that when the input signal represents a chaotic dynamics (in discrete or 
discretized continuous setting)
 RNNs indeed predict chaotic attractor for some 
number of steps and then the predicted trajectories diverge from the actual ones \cite{Vlachas2018,Li2016,Yu2017}.
This result seems natural as 
it reflects an important property of chaotic dynamics --  extremely high sensitivity of chaotic systems to 
small perturbations in initial conditions. 

What does happen when a trajectory is perturbed by 
external noise of specific statistics, e.g., white noise? How would RNN
extrapolate the input of such noisy time series?
Generally speaking, when a noisy signal is used as an input to a \emph{predictive} RNN 
it is expected that a trained network would be able to extrapolate the \emph{noisy}
time series. 
It appeared that the extrapolated trajectory is not noisy --
filtering of the noisy perturbation of the Lorenz attractor dynamics was reported in
\cite{Haykin2001} 
where the authors used recurrent multi-layer perception network and
noted that the reconstructed signals were "reasonably close to the noise-free
signal and the iterated predictions are smoother in comparison to the noisy signals" 
\cite{Haykin2001}.
This observation leads to the following question - given a smooth deterministic 
function with added noise component as a RNN input will the 
trajectory anticipated by RNN be noisy or smooth?
A short note \cite{Yeo2019} considered LSTM network \cite{Hochreiter1997} with 128 neurons trained on the 
Mackey-Glass time series with added noise and demonstrated that 
with the increase of the noise level LSTM  behaviour depends more on its own 
dynamics than on the input data. On the contrary, the training using the 
noiseless input produces RNN with very high sensitivity to small perturbations.

In this manuscript we attempt to explain the fact 
that RNN trained
on segments of \emph{noisy} trajectory and being fed a segment of such trajectory
generates a \emph{smooth} extrapolating curve. Our analysis shows that smooth predictions
are commonplace and independent of the RNN type or extrapolation quality.
We establish conditions for such RNN behavior and 
find that when these conditions are met  a new very fast predictive algorithm
can be implemented. We demonstrate that this algorithm for relatively long
input sequences (around fifty time points) works as good as the original one 
and gives the speed up to an order of magnitude.

The manuscript is organized as follows. Section \ref{algorithm0} describes the architecture 
of a very simple network made of a single recurrent network with small number of neurons 
followed by a linear layer. Section \ref{state_dynamics} describes RNN 
governing transformations and presents a standard algorithm used for
time series prediction. Next Section \ref{training} deals with the network training
and discusses the dependence of the prediction quality on the number
of neurons in RNN.
Section \ref{noise_prop} considers the input noise influence onto RNN state dynamics
and demonstrates that it cannot be neglected.
Then in Section \ref{RNNdynamics} the focus shifts to the RNN dynamics during 
a recursive prediction procedure and conditions when this procedure
results in smooth output are established. We show that satisfaction of these conditions allows
to design a new much faster predictive algorithm described in details in 
Section \ref{algorithm} and demonstrate its high quality 
of extrapolation.
The next Section \ref{neuro} is devoted to possible implications of the 
presented results for neuroscience.
Section \ref{discussion} is devoted to discussion of  possible applications
and generalizations of our findings.

\section{Network architecture and predictive algorithm}
\label{algorithm0}

Consider a simple two layer network 
designed to predict multidimensional time series  $\bm{\mathcal X} = \{\bm x_i\},\ 1\le i \le N$.
The first layer is a recursive network with $n$ neurons -- it takes a subsequence $\bm X_{k,m} = \{\bm x_i\} =
\{\bm x_{k+1},\bm x_{k+2},\ldots,\bm x_{k+m}\},\ 0\le k \le N-m,$ of $m$ vectors $\bm x_i$ having
dimension $d$ each and returns a sequence $\bm S$ of $n$-dimensional state vectors $\bm s_i,\ 
(1 \le i \le m)$. The last element $\bm s_{m}$ is transferred into the 
second linear layer that generates an output vector $\bar{\bm x}$ of dimension $d$ by linear transformation
$\bar {\bm x} = \bm W\cdot\bm s_{m} + \bm b$, with matrix $\bm W$ of dimensions 
$d \times n$ and $d$-dimensional bias vector $\bm b$.

\begin{figure}[h!]
\begin{center}
\psfig{figure=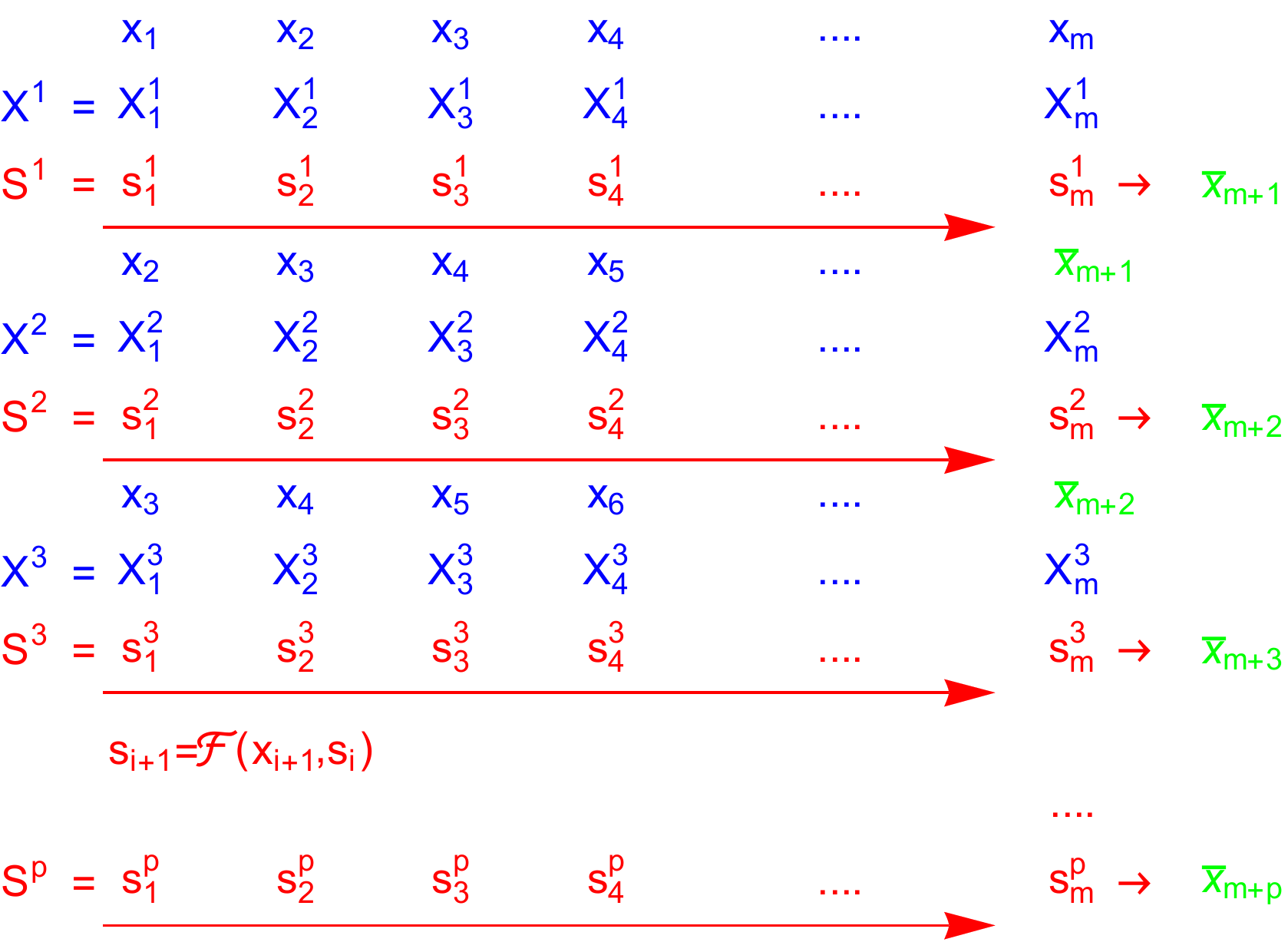,height=8.0cm} 
\caption{The scheme of the prediction double recursive procedure for RNN.
Three first and the last prediction steps are shown. 
The elements of the input sequences $\bm X^j$ to RNN (blue) are fed into
(\ref{map0}) 
 to produce recursively recurrent network states $\bm s^j_i$ (red). The last element $\bm s^j_m$
in $\bm S^j$ is transformed by
(\ref{linear}) to generate the predicted point $\bar{\bm x}_{m+j+1}$ (shown in green).
This point is used to update the input sequence $\bm X^{j+1}$ for the 
next prediction step.
}
\label{Fig0}
\end{center}
\end{figure}

A trained network is used for time series prediction recursively.
Namely, one starts with a sequence $\bm X^{1} =\bm X_{k,m}$ of length $m$ supplied as input
to the RNN; the resulting output is considered as a prediction of the next time point  
$\bar {\bm x}_{k+m+1}$
of the input sequence. The next input sequence $\bm X^{2}$ to RNN 
is produced by dropping the first point of $\bm X^{1}$ and 
adding the predicted point to the result:
$\bm X^{2} = \bm X_{k+1,m-1} \cup \bar {\bm X}_{k+m,1} $;
here $\cup$ denotes union of two sequences with order of elements preserved.
This sequence is used as input to the RNN that generates $\bar{\bm x}_{k+m+2}$
and a next input ${\bm X}^{3} = {\bm X}_{k+2,m-2} \cup \bar {\bm X}_{k+m,2} $ is formed.
Thus at $j$-th predictive step ($j \le m$) the input $\bm X_k^j$ to RNN is 
formed as ${\bm X}^{j} = {\bm X}_{k+j-1,m-j+1} \cup \bar {\bm X}_{k+m,j-1}$,
while for $j > m$ the input is formed by the already predicted values only
${\bm X}^{j} = \bar {\bm X}_{k+j-m-1,m}$.
The recursive procedure is repeated $p$ times to produce $p$ new time points
$\bar{\bm x}_{k+m+i}, (1 \le i \le p)$ approximating the time series $\bm{\mathcal X}$
segment $\{{\bm x}_i\}$ for $k+m+1 \le i \le k+m+p$ (Figure \ref{Fig0}).
As the offset value $k$ determining the initial point
of the input sequence $\bm X^1$ is arbitrary but fixed 
for given predictive procedure, 
without loss of 
generality we further set it equal to zero.
The described algorithm can be called a {\it moving window} prediction as it
is characterized by recurrent usage of the input sequence 
$\bm X^j$ obtained  from $\bm X^{j-1}$  by shifting one position to the right. 
It is easy to see that the procedure uses a double recursion --
the inner one used $m$ times in the recurrent layer and the outer is employed $p$ 
times to generate the output points, so that the total number of recursions is 
$mp$.



\section{Network state dynamics}
\label{state_dynamics}

In this manuscript  we perform the analysis of all standard recurrent networks
-- basic, gated  and LSTM RNNs.
Consider an inner dynamics of a recurrent network in more details.
The input sequence $\bm X = \{\bm  x_i\},1 \le i \le m$ produces the
network state sequence $\bm S = \{\bm s_i\}$ for the basic network
\be
\bm s_i = \tanh (\bm W_{ix} \cdot \bm x_{i} + \bm W_{is} \cdot \bm s_{i-1} + \bm b_i),
\label{basic}
\ee
where $\bm W_{ix} ,\ \bm W_{is} $ are matrices and $\bm b_i$ is a bias vector.
The gated network \cite{Chung2014} is governed by the following relations
 \bea
\bm i_i &=& \sigma(\bm W_{ix} \cdot \bm x_{i} + \bm W_{is} \cdot \bm s_{i-1} + \bm b_i),
\nonumber \\
\bm r_i &=& \sigma(\bm W_{rx} \cdot \bm x_{i} + \bm W_{rs} \cdot \bm s_{i-1} + \bm b_r),
\nonumber \\
\bm m_i &=& \tanh(\bm W_{mx} \cdot \bm x_{i} + \bm r_i \otimes \bm W_{ms} \cdot \bm s_{i-1} + \bm b_m),
\nonumber \\
\bm s_i &=& (1 - \bm i_i) \otimes \bm m_i +   \bm i_i \otimes \bm s_{i-1},
\quad
\bm a \otimes \bm b  = \sum_{k} a_k b_k,
\label{Fgated}
\eea
where $\sigma(x) = 1/(1+\exp(-x))$ is the logistic sigmoid function,
$\otimes$ denotes the elementwise multiplication of two vectors of the same length
and the initial state $\bm s_0 = \bm 0$. The vectors $\bm i_i, \ \bm r_i, \ \bm m_i$ denote
the input, reset and memory gate state respectively.

For LSTM network \cite{Hochreiter1997} the governing transformation that determines
network state $\bm S = \{\bm s_i\}$ and cell state 
$\bm C = \{\bm c_i\}$ sequences
is defined by
\bea
\bm s_i & = & \bm o_i \otimes \tanh \bm c_i,
\nonumber \\
\bm c_i & = & \bm f_i \otimes \bm c_{i-1} + \bm i_i \otimes \bm m_{i},
\nonumber \\
\bm o_i & = & \sigma( \bm W_{ox}\bm x_i +\bm W_{os}\bm s_{i-1} + \bm b_o ),
\nonumber \\
\bm i_i & = & \sigma( \bm W_{ix}{\bm x}_i + \bm W_{is}\bm s_{i-1} + \bm b_i ),
\label{LSTM} \\
\bm f_i & = & \sigma( \bm W_{fx}\bm x_i + \bm W_{fs}\bm s_{i-1} + \bm b_f ),
\nonumber \\
\bm m_i & = & \tanh( \bm W_{mx}\bm x_i + \bm W_{ms}\bm s_{i-1} +\bm b_m),
\nonumber
\eea
where 
the initialization value of state $\bm s_0$ and 
cell state $\bm c_0$  vector is zero vector of length $n$.
With $a=i,f,m,o$ we denote $\bm W_{ax}, \ \bm W_{as}$  matrices and $\bm b_{a}$ bias
vectors for the  input, forget, memory and output gates respectively; all these structures are trainable
and in the trained network their elements are real valued constants.

The shorthand form of the transformations (\ref{basic}-\ref{LSTM}) reads
$
{\bm s}_i = \bm{\mathcal F}({\bm x}_i, {\bm s}_{i-1},{\bm P}),
$
where $\bm P$ denotes elements of all matrices and bias vectors in (\ref{basic}-\ref{LSTM}) and
$\bm s_0$ is $n$-dimensional zero vector. As in trained network the set $\bm P$ is fixed we will drop it from the 
list of arguments of the vector function $\bm{\mathcal F}$
\be
{\bm s}_i = \bm{\mathcal F}({\bm x}_i, {\bm s}_{i-1}).
\label{map0}
\ee
It is important to note that the governing transformations imply for every step $i$  in (\ref{map0})
all components of $\bm s$ satisfy a condition
$|s_{k}| \le 1, 1 \le k \le n$.
The equations (\ref{basic}-\ref{LSTM})  are accompanied by a linear transformation
\be
\bar{\bm x}_{m+1} = \bm W \cdot \bm s_m + \bm b,
\label{linear}
\ee
where $\bar{\bm x}_{m+1}$ is a value predicted by RNN based on the input $\bm X$.

\section{RNN training and performance}
\label{training}
The RNNs we use in the simulation have a small number $n$ of neurons in the recurrent layer
$1 \le n \le 20$. The training set is constructed by merging $6000$  segments of 
variable length ($5 \le m \le 150$)  of two periodic 
one-dimensional ($d=1$) functions -- the sine wave $g_0(t) = \sin(2\pi t)$ and 
the shifted triangle wave
$h_0(t)  = 1/2 + 1/\pi \arcsin (\sin 2\pi x)$. The white noise with the amplitude
$a=0.15$ is added to both functions -- 
$g(t) = g_0(t) + a\xi(t), \ h(t) = h_0(t) + a\xi(t)$.
The time step $\Delta t$ between the adjacent time points is selected equal to
$\Delta t = 0.01$. 
The RNNs are trained for 50 epochs on the complete set of $12000$
segments with $20\%$ validation set using Adam algorithm.
The RNNs fail to predict the noisy dynamics of $g(t)$ or $h(t)$, instead
all RNNs produce some smooth predictions $G_0(t)$ and $H_0(t)$, respectively.
We define the quality function of prediction $F(t)$ 
vs. the actual dynamics $f(t)$   ($f = g,h$ and $F = G,H$)
as 
$$
Q^{-1} = \frac{1}{p}\sum_{i=1}^p \lVert F(t_i) - f(t_i) \rVert^2,
$$
where $p$ is the length of the predicted sequence and $\lVert\ \rVert$ denotes the Euclidean norm.

\begin{figure}[h!]
\begin{center}
\begin{tabular}{ccc}
{\bf a} &
\psfig{figure=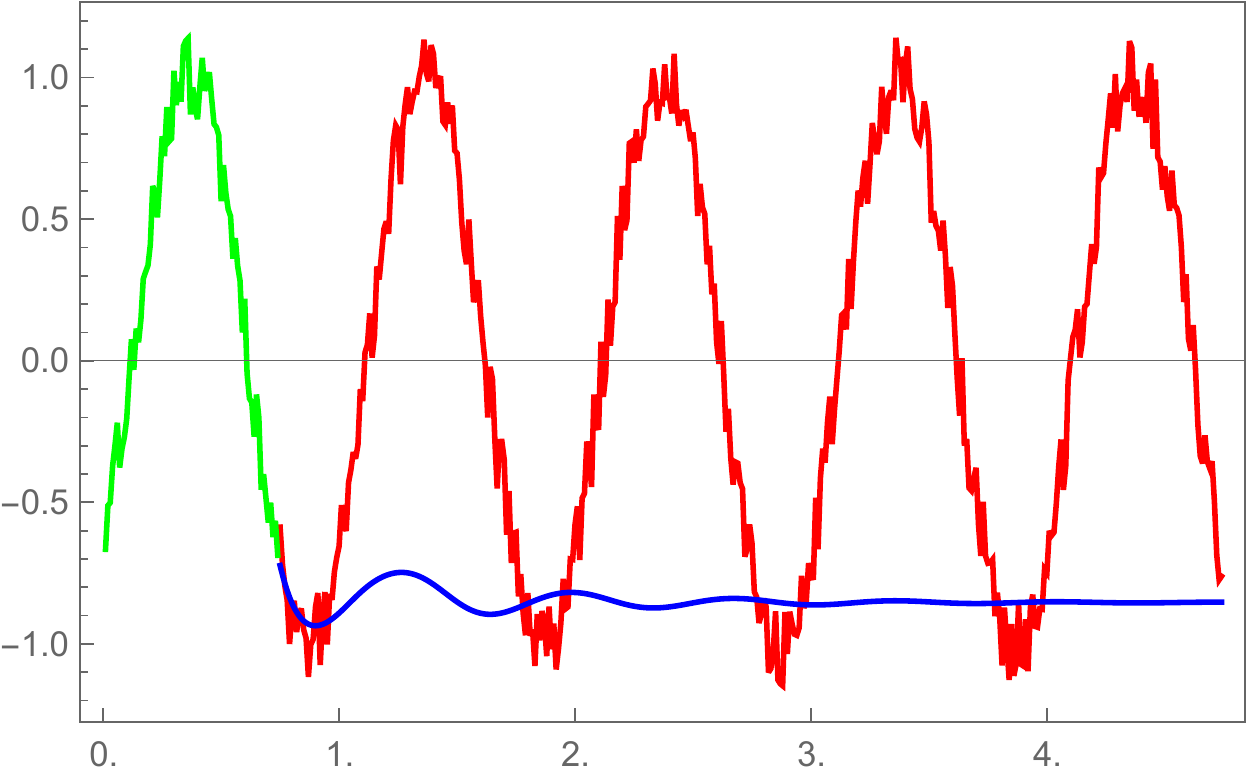,height=4.0cm} 
&
\psfig{figure=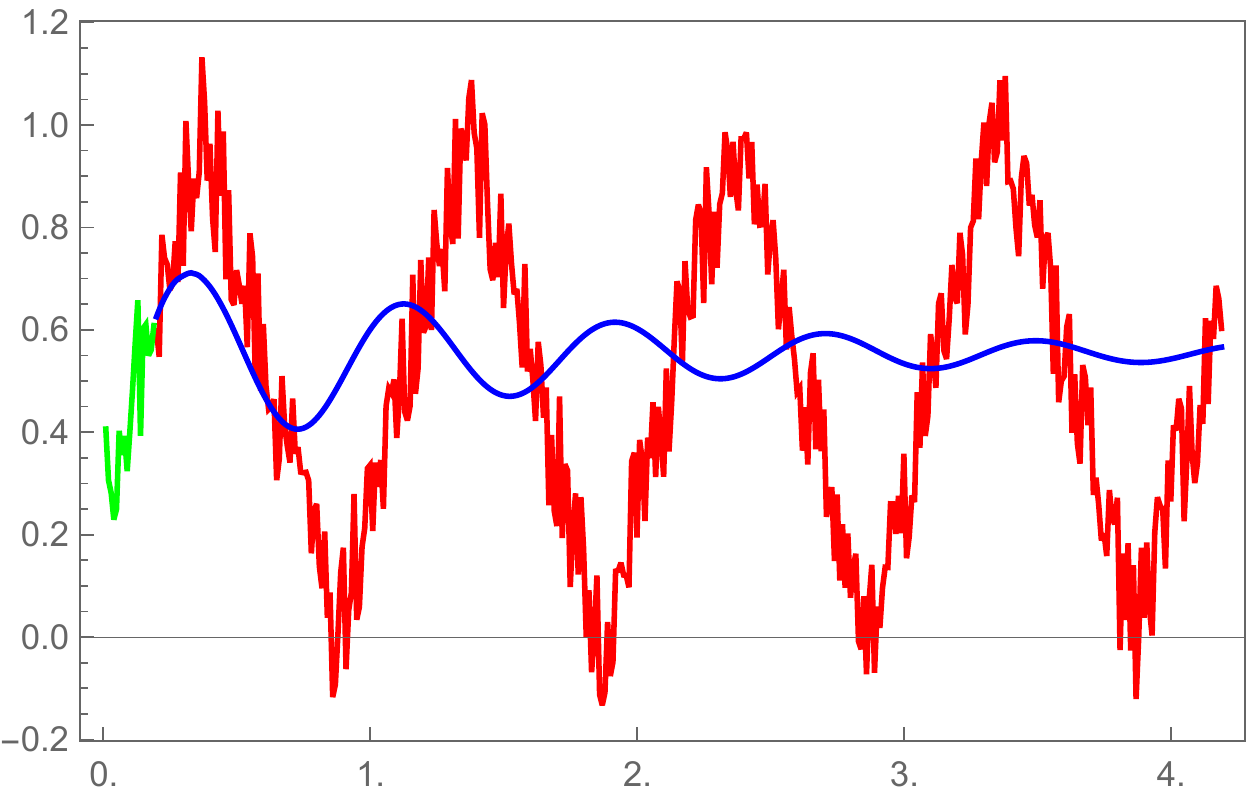,height=4.0cm}
\\
{\bf b} &
\psfig{figure=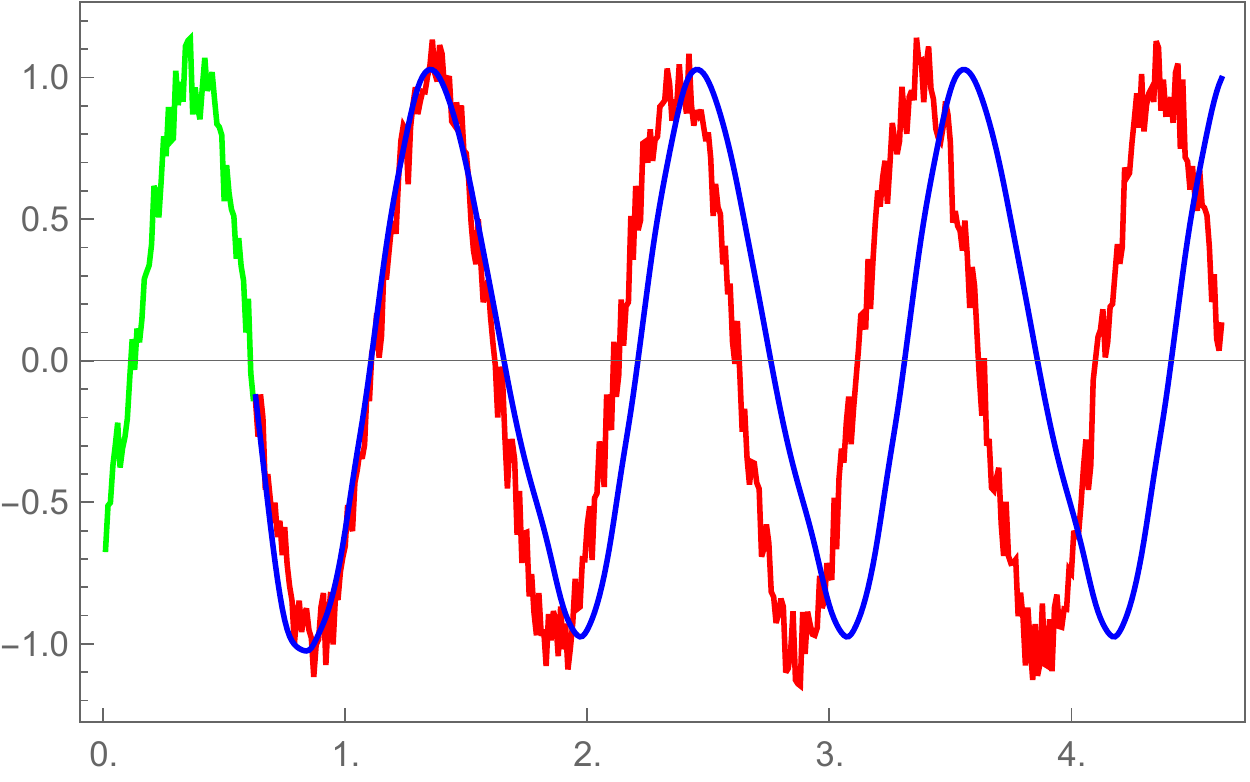,height=4.0cm} 
&
\psfig{figure=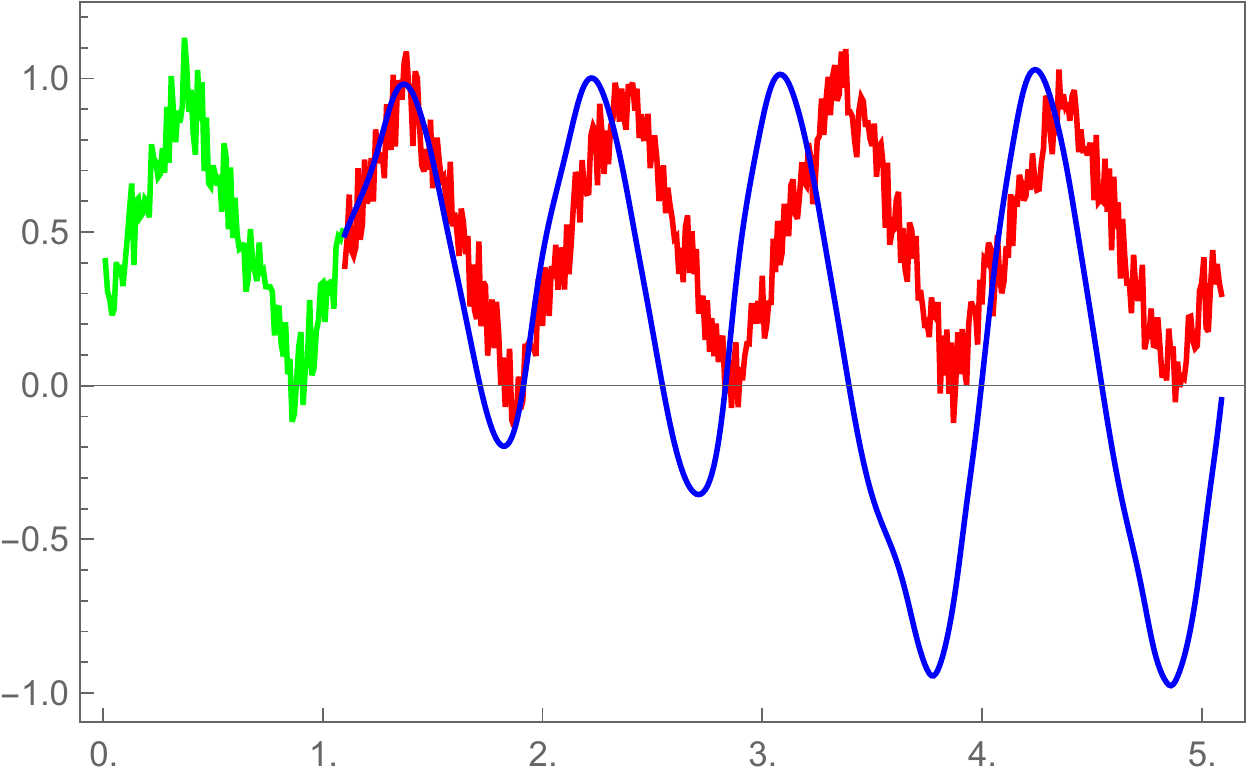,height=4.0cm}
\\
{\bf c} &
\psfig{figure=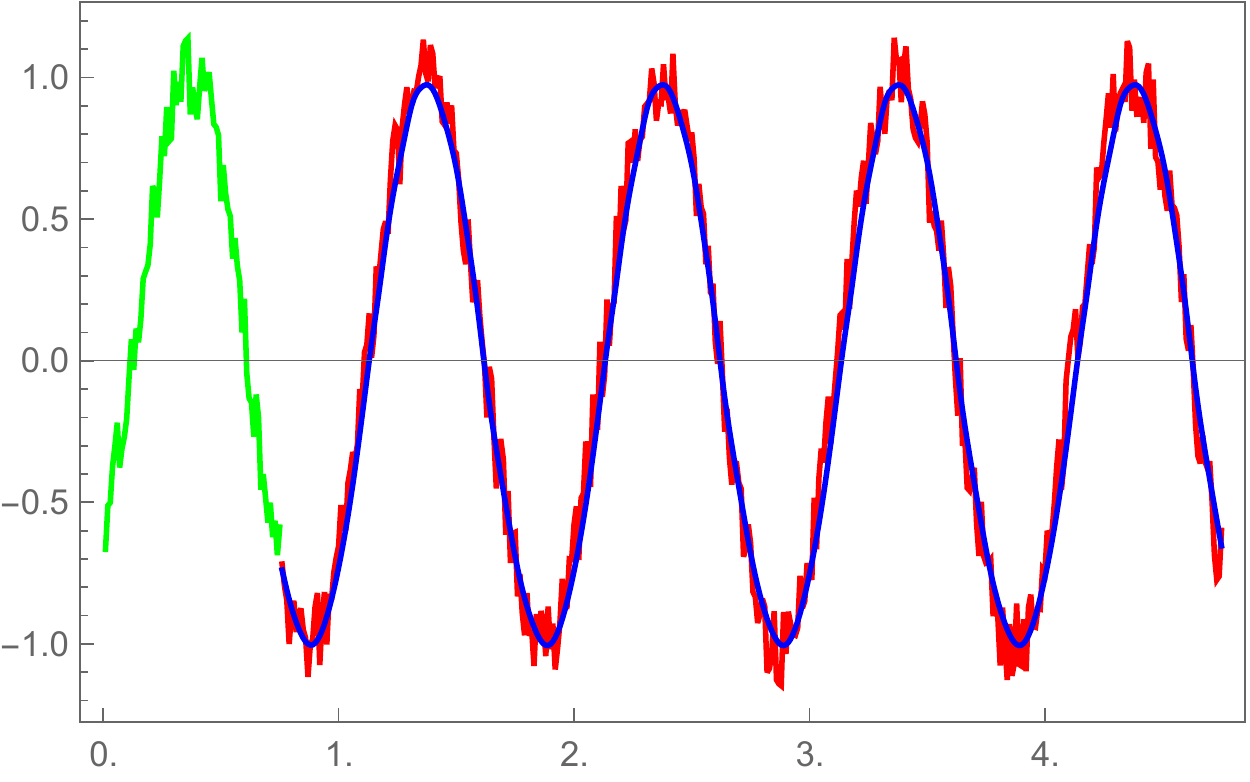,height=4.0cm} 
&
\psfig{figure=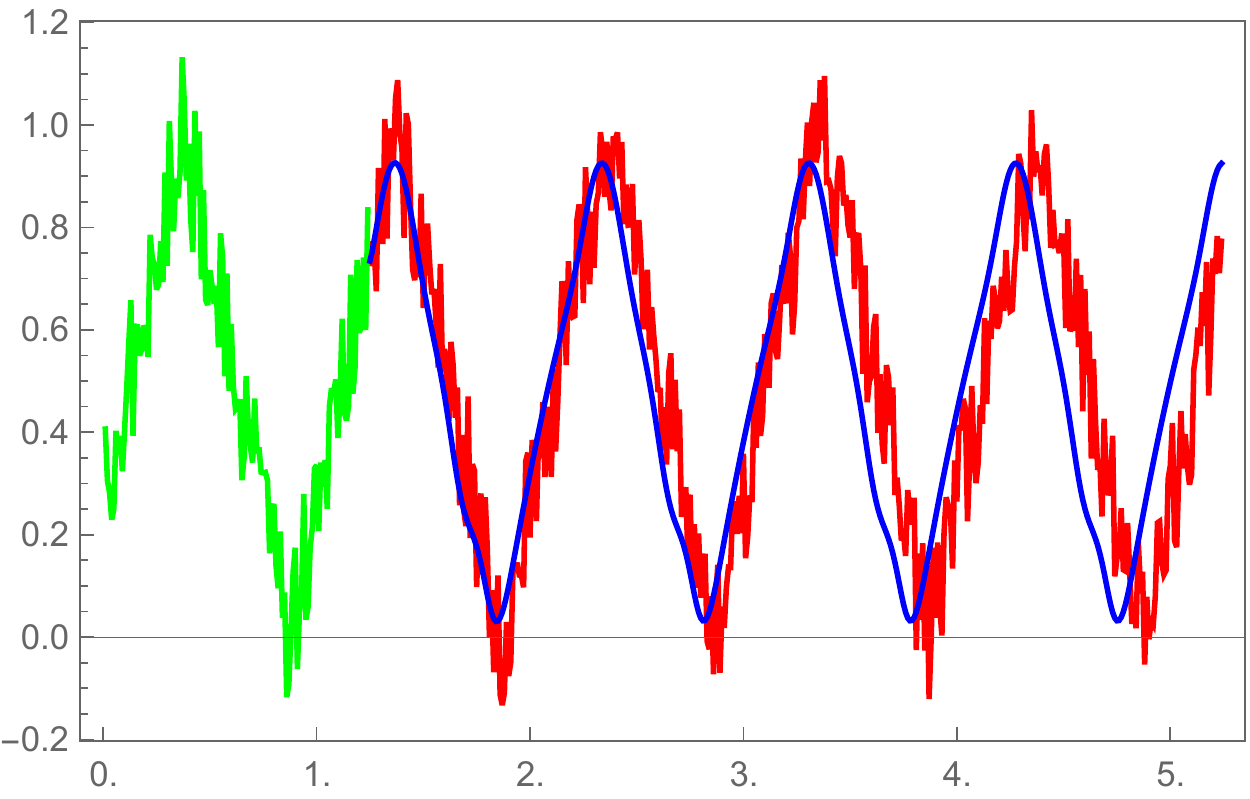,height=4.0cm}
\end{tabular}
\caption{The input segment of the noisy ($a=0.15$) sequence (green) of 
sine (left) and triangle (right) waves, the subsequent segment of 
$\bm{\mathcal X}$
(red) and predicted dynamics (blue) for 
({\bf a}) $5$, ({\bf b}) $10$, ({\bf c}) $20$  neurons in LSTM network.
}
\label{Fig1}
\end{center}
\end{figure}

As it was expected 
the value of $Q$ for the LSTM network increases with $n$ (see  Figure \ref{Fig1}). Nevertheless the predicted dynamics
is always smooth which implies that the filtering property of  RNN is 
independent of the prediction quality. We observe that for $n=10$ the deviation of the 
predicted curve from the actual one is quite small for one period ($Q > 30$).
Note that the prediction of the underlying smooth function was very good
for $n=20$ neurons ($Q > 100$) which is much smaller than ($n=128$) reported in \cite{Yeo2019}.
Qualitatively similar results are obtained for the basic recurrent network.

\section{Noise propagation in recurrent network}
\label{noise_prop}

Consider the process of state vector computation assuming that the input sequence  ${\bm X}$
represents time point values of the function $\bm g(t) =\bm g_0(t) + a \bm{\xi}(t)$ 
where $\bm g_0(t)$ is a smooth function, $\bm{\xi}$ is a white noise random process with a small amplitude
 $0 \le a \ll 1$. This implies that RNN is trained to predict the 
values ${\bm x}_i = \bm g_0(t_i) + a \bm{\xi}(t_i)$ for $i > m$ using the input  
${\bm X}_{m}$. As the parameters of the state transformations are constants
one expects that the values $\bm s_i$ for $i>0$ might contain a noisy component and that
eventually a sequence $\bar{\bm X}_{m,p}$ of the predicted values would
be a representation of some noisy function.
In other words, RNN is expected to produce a discrete representation 
of a function $\bm G(t)$ that mimics with some accuracy
the {\it noisy} function $\bm g(t)$  using 
the {\it noisy} input ${\bm X}_{m}$ representing the same function $\bm g(t)$.

Consider step by step computation of $\bm s_i$. Using smallness of the noise amplitude
$a$ we find for $\bm s_1$ from (\ref{map0}) using Taylor expansion in $a$
in linear approximation
\be
\bm s_1 = \bm{\mathcal F}(\bm g_0(t_1) + a\bm \xi_1,\bm 0) \approx
\bm{\mathcal F}(\bm g_0(t_1),\bm 0) + 
a \bm{\mathcal F}'(\bm g_0(t_1),\bm 0) \otimes \bm \eta_1 = 
\hat{\bm s}_1 + a \tilde{\bm s}_1 \otimes \bm\eta_1,
\label{step1}
\ee
where $\bm \eta$ is a $n$-dimensional random process obtained
by a linear transformation of the $d$-dimensional random process $\bm \xi$.
The computation of $\bm s_2$ gives
\bea
\bm s_2 &=& \bm{\mathcal F}(\bm g_0(t_2) + a\bm \xi_2,\hat{\bm s}_1 + a \tilde{\bm s}_1 \otimes \bm\eta_1) 
\nonumber \\
&\approx&
\bm{\mathcal F}(\bm g_0(t_2),\hat{\bm s}_1) + 
a \bm{\mathcal F}'(\bm g_0(t_2),\hat{\bm s}_1) \otimes
(\bm \eta_2 +\bar{\bm W}\cdot \tilde{\bm s}_1 \otimes \bm\eta_1) 
\nonumber\\
&=& \bm{\mathcal F}(\bm g_0(t_2),\hat{\bm s}_1) + 
a \bm{\mathcal F}'(\bm g_0(t_2),\hat{\bm s}_1) \otimes \bm\zeta_2 = 
\hat{\bm s}_2 + a \tilde{\bm s}_2 \otimes \bm\zeta_2,
\label{step2}\\
&& 
\bm\zeta_2 = \bm \eta_2 +\bar{\bm W}\cdot \tilde{\bm s}_1 \otimes \bm\eta_1,
\nonumber
\eea
where $\bar{\bm W}$ denotes a matrix used in transformation of the noise
component generated in the vector $\bm s_1$. 

The subsequent steps ($1 \le k \le m$) produce 
\bea
\bm s_k &=& \bm{\mathcal F}(\bm g_0(t_k) + a\bm \xi_k,\hat{\bm s}_{k-1} + a \bar{\bm s}_{k-1} \otimes \bm\zeta_{k-1})
\nonumber \\
&\approx&
\bm{\mathcal F}(\bm g_0(t_k),\hat{\bm s}_{k-1}) + 
a \bm{\mathcal F}'(\bm g_0(t_k),\hat{\bm s}_{k-1}) \otimes
(\bm \eta_k +\bar{\bm W}\cdot \tilde{\bm s}_{k-1} \otimes \bm\zeta_{k-1}) 
\nonumber\\
&=& \bm{\mathcal F}(\bm g_0(t_k),\hat{\bm s}_{k-1}) + 
a \bm{\mathcal F}'(\bm g_0(t_k),\hat{\bm s}_{k-1}) \otimes \bm\zeta_k 
= \hat{\bm s}_k + a \tilde{\bm s}_{k} \otimes \bm\zeta_k,
\label{stepk}
\\
&& 
\bm\zeta_k = \bm \eta_k +\bar{\bm W}\cdot \tilde{\bm s}_{k-1} \otimes \bm\zeta_{k-1},
\nonumber
\eea
where 
$$
\hat{\bm s}_k = \bm{\mathcal F}(\bm g_0(t_k),\hat{\bm s}_{k-1}),
\quad
\tilde{\bm s}_{k} = \bm{\mathcal F}'(\bm g_0(t_k),\hat{\bm s}_{k-1}) ,
$$
and the derivative is taken w.r.t. noise amplitude $a$.
Note that (\ref{stepk}) is valid for $k=1,2$ if one defines 
$\bm\zeta_1 = \bm\eta_1 + \tilde{\bm s}_{0} \otimes \bm\zeta_{0}$, and
$\tilde{\bm s}_{0}$ as zero vector.

From (\ref{stepk}) it follows that the last element $\bm s_m$ of the state sequence also has the
noise contribution $a \tilde{\bm s}_{m} \otimes \bm\zeta_m$ which 
naturally transfers to the first predicted value 
$$
\bar{\bm x}_{m+1} = 
\bm W\cdot\hat{\bm s}_m+\bm b + a\bm W\cdot\tilde{\bm s}_{m} \otimes \bm\zeta_m
= \bm G(t_{m+1}) = \bm G_0(t_{m+1}) +  a\bm W\cdot\tilde{\bm s}_{m} \otimes \bm\zeta_m,
$$
where $\bm G$ and $\bm G_0$ are approximations to the functions 
$\bm g$ and $\bm g_0$ generated by RNN.
This means that the RNN itself only transforms the input noise but does not filter it out.

The predicted element $\bar{\bm x}_{m+1}$ is used  as the last element of the input sequence 
in the next prediction step and therefore one expects that the predicted sequence 
$\bar{\bm X}_{m,p}$ should reflect the noise components contained both in the input and 
predicted sequences.
Unexpectedly, the numerical experiments (see below) show that in fact the predicted sequence
$\bar{\bm X}_{m,p}$ is not noisy but represents the approximation $\bm G_0(t)$ of the smooth
function $\bm g_0(t)$. The goal of this manuscript is to explain this unexpected behavior 
and to determine conditions required for generation of a smooth prediction.



\section{RNN state dynamics}
\label{RNNdynamics}

In the previous Section we observe that the noise component of the input signal is preserved in the 
RNN states, and we have to look at state dynamics in more details to 
understand noise filtering in the trajectory prediction process.

\subsection{Numerical experiments}
\begin{figure}[h!]
\begin{center}
\begin{tabular}{cc}
\psfig{figure=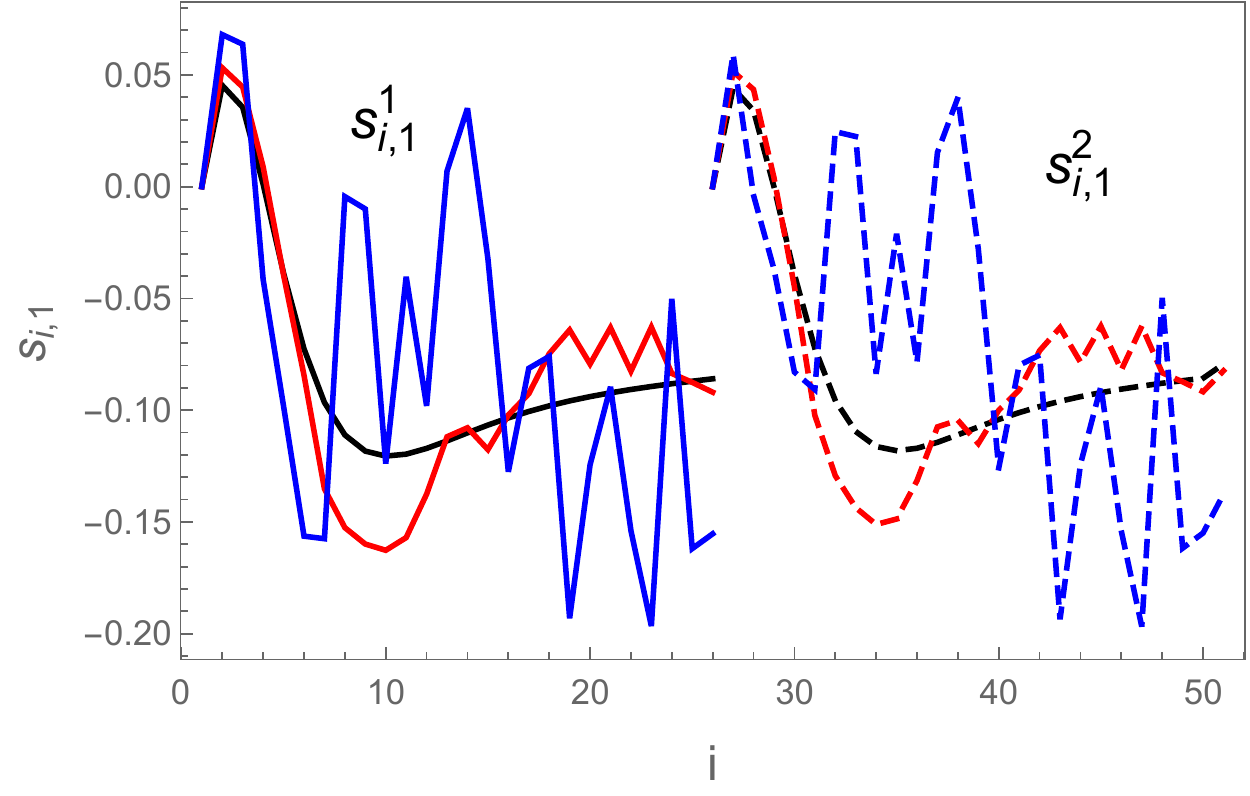,height=5.0cm}  &
\psfig{figure=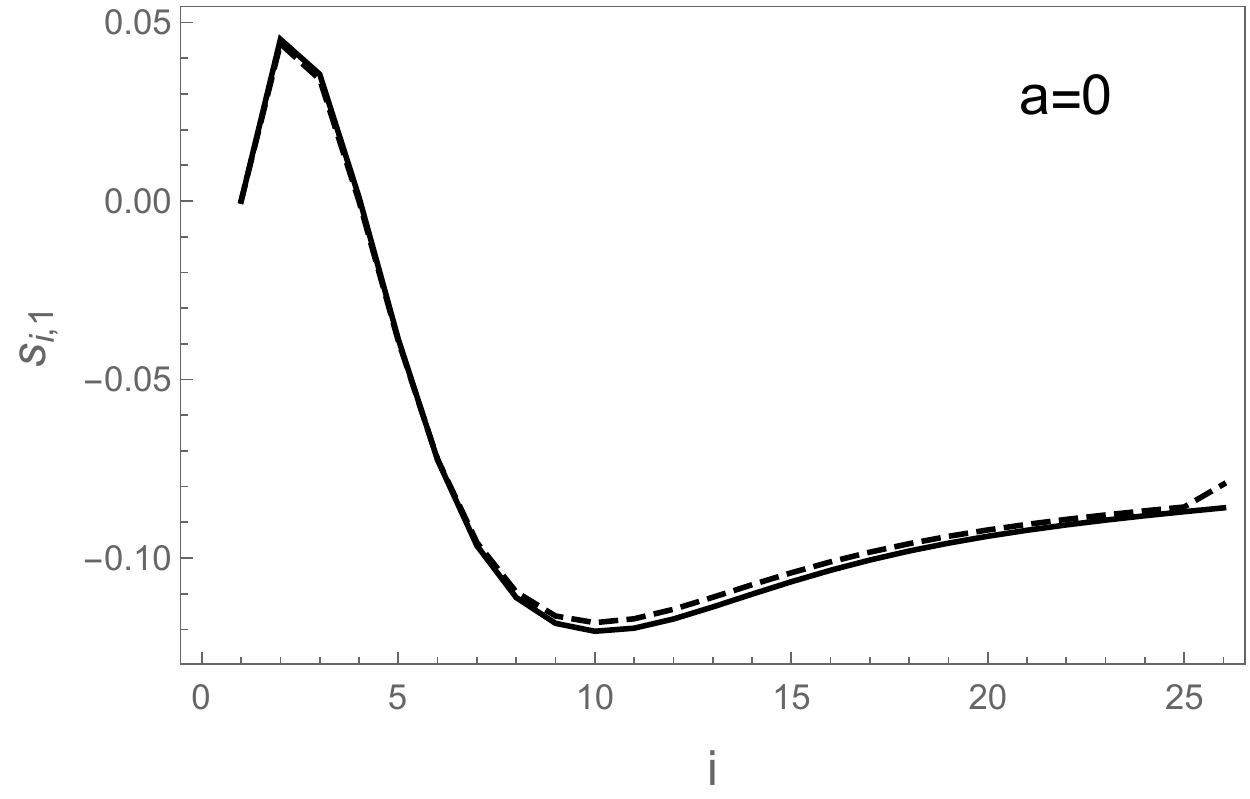,height=5.0cm}
\\
{\bf a} & {\bf b} \\
\psfig{figure=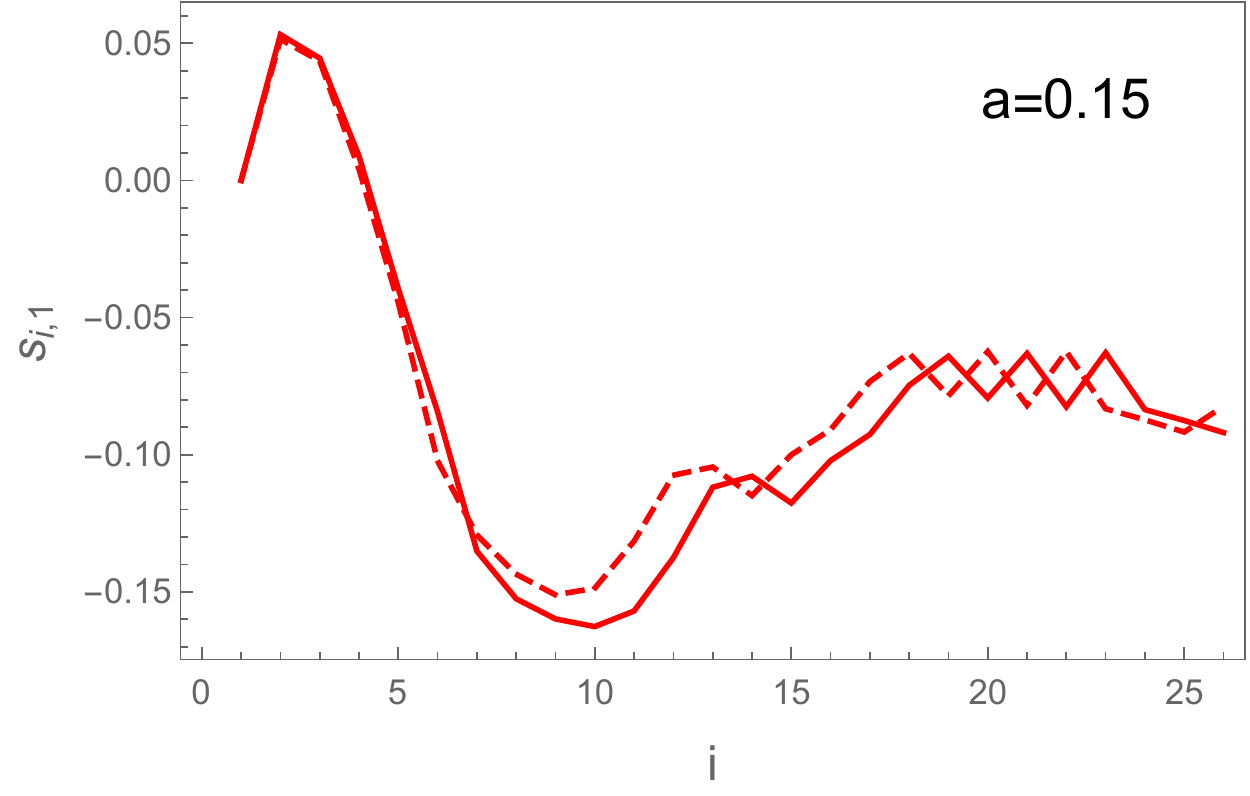,height=5.0cm} &
\psfig{figure=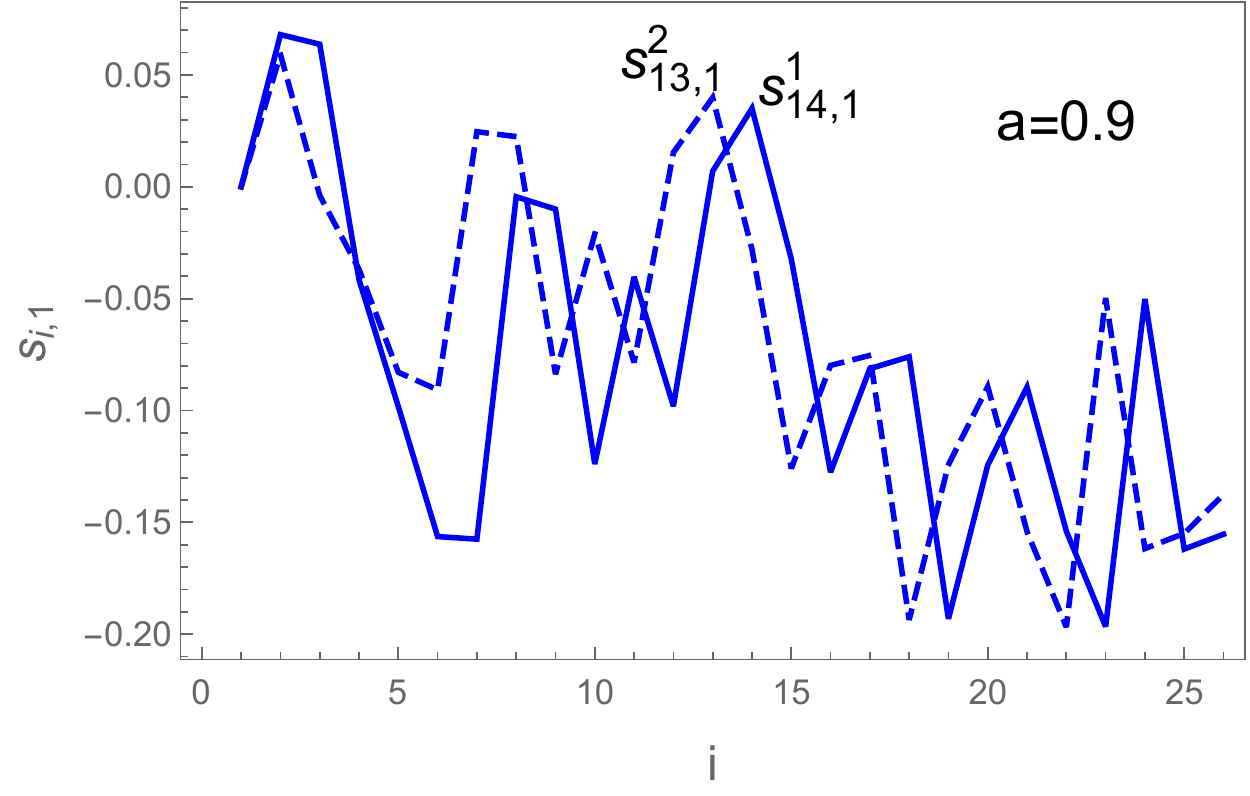,height=5.0cm}
\\
{\bf c} & {\bf d}
\end{tabular}
\caption{The dynamics of the first element $s^j_{i,1}$ of the state vector 
$\bm s^j_i$ in the $j$-th round of prediction for $j=1$ (solid) and
$j=2$ (dashed) for three noise amplitudes -- 
$a=0$ (black), $a=0.15$ (red) and $a=0.9$ (blue).
({\bf a}) The sequence $s^2_{i,1}$ is shifted w.r.t. of $s^1_{i,1}$.
({\bf b} - {\bf d}) The sequences are overlapped for different noise amplitudes: 
({\bf b}) $a=0$ (no noise),
({\bf c}) original amplitude $a=0.15$,
({\bf d}) increased amplitude $a=0.9$. 
The values of $s^1_{i+1,1}$ and $s^2_{i,1}$ tend to each other
with increasing $i$.
}
\label{Fig2}
\end{center}
\end{figure}

Consider in details the sequence of the RNN states $\bm S^1$ and $\bm S^2$
for first and second prediction steps for three values of the noise amplitude 
$a=0, 0.15, 0.9$  of the input sequence. Figure \ref{Fig2}a 
demonstrates that indeed the dynamics of LSTM state is affected by 
noise as predicted by (\ref{stepk}). We also note that both sequences $\bm S^1$ and $\bm S^2$ look very similar.
To test this similarity we overlay the corresponding sequences for 
given noise amplitude (Figure \ref{Fig2}b-d) and find that even in case of large noise $a=0.9$
the sequence $\bm S^2$ is very close to the sequence $\bm S^1$ shifted by one step to the left,
in other words $\bm s^2_{i} \approx \bm s^1_{i+1}$.

\subsection{Dynamics of state vector shifted difference}
To understand this behavior recall a relation between the input sequences 
$\bm X^j$ and $\bm X^{j+1}$ (see Figure \ref{Fig0}).
The input sequence $\bm X^{j}$ construction algorithm described in 
Section \ref{state_dynamics} implies that
$\bm X^{j+1}_i=\bm X^{j}_{i+1}$ for all $2\le i \le m-1$.
Using (\ref{map0}) we find
\bea
{\bm s}^1_{i+1}&=&\bm{\mathcal F}({\bm  X}^1_{i+1}, {\bm s}^1_{i}),
\quad \quad \quad  \quad \quad 0 \le i \le m-1,
\label{pred0_i+1}
\\
{\bm s}^2_i &=& \bm{\mathcal F}({\bm  X}^2_i, {\bm s}^2_{i-1}) = 
\bm{\mathcal F}({\bm  X}^1_{i+1}, {\bm s}^2_{i-1}),
\quad 1\le i \le m-1.
\label{pred1_i}
\eea
We observe that in computation of ${\bm s}^1_{i+1}$ and 
${\bm s}^2_{i}$ the first argument of the map $\bm{\mathcal F}$
in  (\ref{pred0_i+1},\ref{pred1_i}) is the same. 
Consider the difference $\bm \delta_i^{1} = {\bm s}^1_{i+1}-{\bm s}^2_{i}$.
For $i=0$ we have  $\bm \delta_1^{1} = {\bm s}^1_{1}=
\bm{\mathcal F}({\bm  X}^1_{1}, \bm 0)$.
For $i=1$ find
$$
\bm \delta_1^{1} = {\bm s}^1_{2}-{\bm s}^2_{1} = 
\bm{\mathcal F}({\bm X}^1_{2}, {\bm s}^1_{1}) -
\bm{\mathcal F}({\bm  X}^1_{2}, \bm 0) =
\bm{\mathcal F}({\bm  X}^1_{2},\bm \delta_0^{1}) -
\bm{\mathcal F}({\bm  X}^1_{2}, \bm 0).
$$
Assuming $\lVert \bm \delta_0^{1} \rVert \ll 1$ expand the first term
above and retain the leading order to obtain
\be
\bm \delta_1^{1} = \frac{\partial \bm{\mathcal F}({\bm  X}^1_{2}, \bm s = \bm 0)}{\partial \bm s} \cdot
\bm \delta_0^{1} = \bm A^1_1\cdot \bm \delta_0^{1}.
\label{delta_1} 
\ee
With $i=2$ find
$$
\bm \delta_2^{1} = {\bm s}^1_{3}-{\bm s}^2_{2} = 
\bm{\mathcal F}({\bm  X}^1_{3}, {\bm s}^1_{2}) -
\bm{\mathcal F}({\bm  X}^1_{3}, {\bm s}^2_{1}) =
\bm{\mathcal F}({\bm  X}^1_{3},{\bm s}^2_{1}+\bm \delta_1^{1}) -
\bm{\mathcal F}({\bm  X}^1_{3},  {\bm s}^2_{1}),
$$
and the expansion leads to
\be
\bm \delta_2^{1} = \frac{\partial \bm{\mathcal F}({\bm  X}^1_{3}, \bm s = {\bm s}^2_{1})}{\partial \bm s} \cdot
\bm \delta_1^{1} = \bm A^1_2\cdot \bm \delta_1^{1} =\bm A^1_1\cdot\bm A^1_2\cdot \bm \delta_0^{1}.
\label{delta_2} 
\ee
It is easy to deduce that for $i=m-1$
\be
\bm \delta_{m-1}^{1} = \bm A^1 \cdot \bm \delta_0^{1},
\quad
\bm A^1 = \prod_{k=0}^{m-1} \bm A^1_k,
\quad 
\bm A^1_k =  \frac{\partial \bm{\mathcal F}({\bm  X}^1_{k+1}, \bm s = {\bm s}^2_{k-1})}{\partial \bm s}.
\label{delta_i} 
\ee
Generalizing the above relations to the other rounds of the predictive cycle we obtain for
$\bm \delta_{m-1}^{j} = {\bm s}^{j}_{m}-{\bm s}^{j+1}_{m-1}$:
\be
\bm \delta_{m-1}^{j} = \bm A^j \cdot \bm \delta_0^{j},
\quad
\bm A^j = \prod_{k=0}^{m-1} \bm A^{j}_k,
\quad 
1 \le j \le p.
\label{delta_i_gen} 
\ee
The numerical simulations of the state dynamics in
the basic and gated RNNs demonstrate the exponential decay of 
shifted difference norm (Figure \ref{FigB1}a,b).
In Appendix for the basic RNN we find the 
approximate explicit expression of the matrix $\bm A_k^j$ and
discuss requirements for the exponential decay of 
$\delta^{j}_i$.
\begin{figure}[h!]
\begin{center}
\begin{tabular}{cc}
\psfig{figure=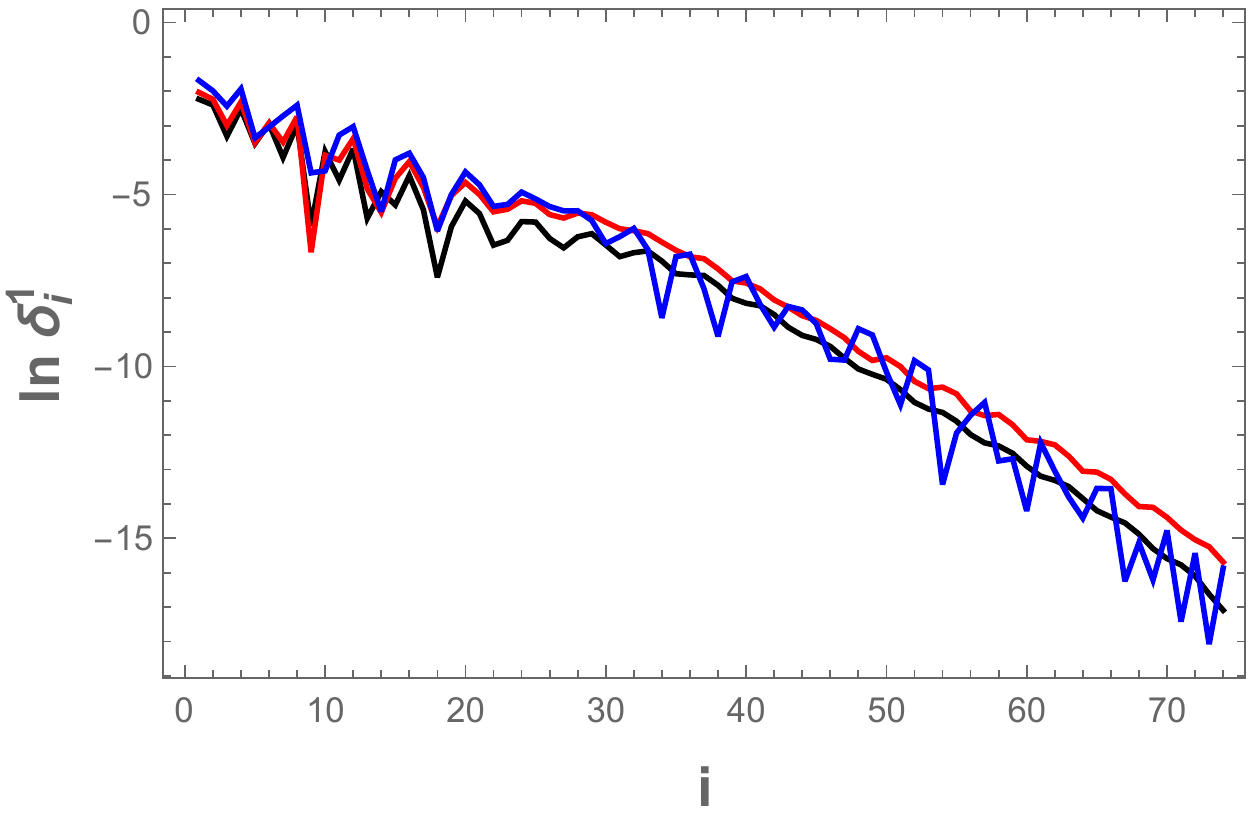,height=5.0cm} &
\psfig{figure=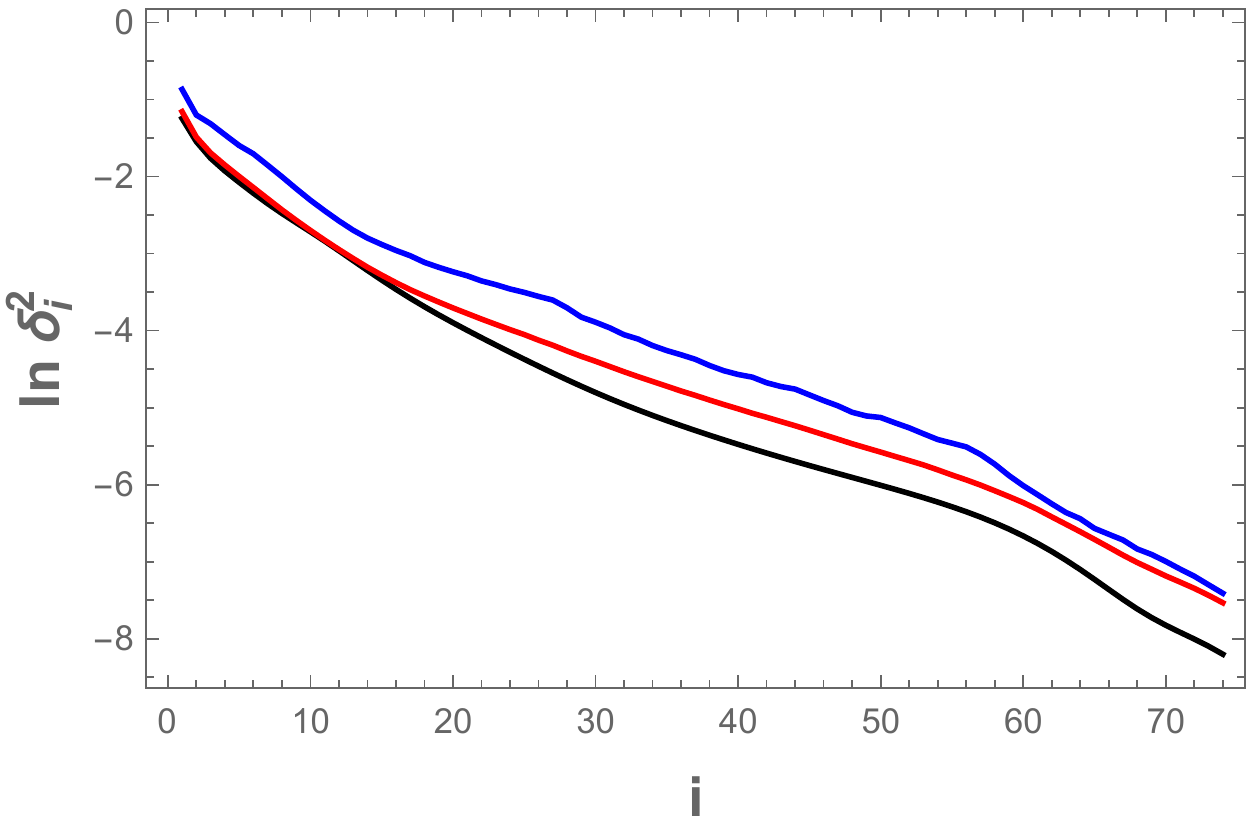,height=5.0cm}
\\
{\bf a} & {\bf b}
\end{tabular}
\caption{The dynamics of the shifted difference norm  
({\bf a}) $\delta^{1}_i$  in the basic RNN with $n=10$,
({\bf b}) $\delta^{2}_i$  in the gated RNN with $n=20$,  
for the noise amplitude
$a=0$ (black), $0.15$ (red) and $0.9$ (blue).}
\label{FigB1}
\end{center}
\end{figure}

\begin{figure}[h!]
\begin{center}
\begin{tabular}{cc}
\psfig{figure=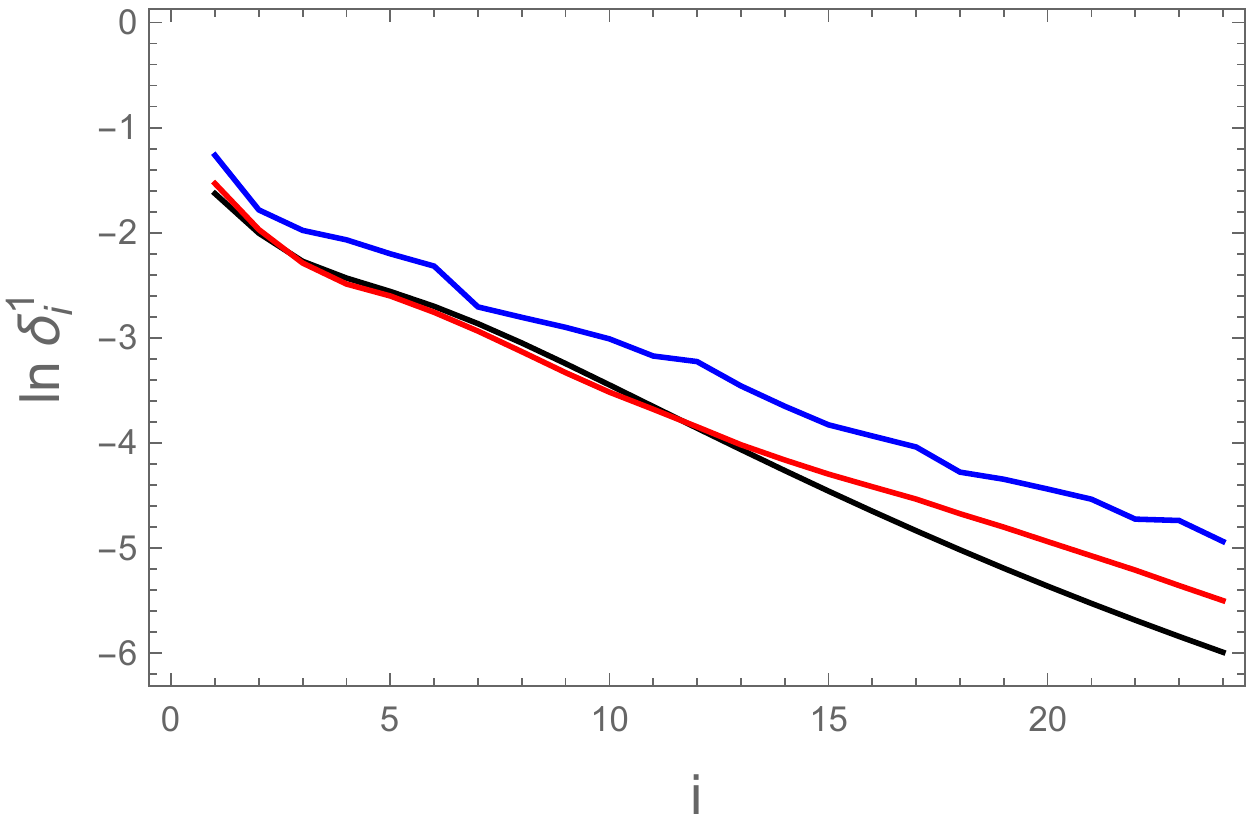,height=5.0cm}  &
\psfig{figure=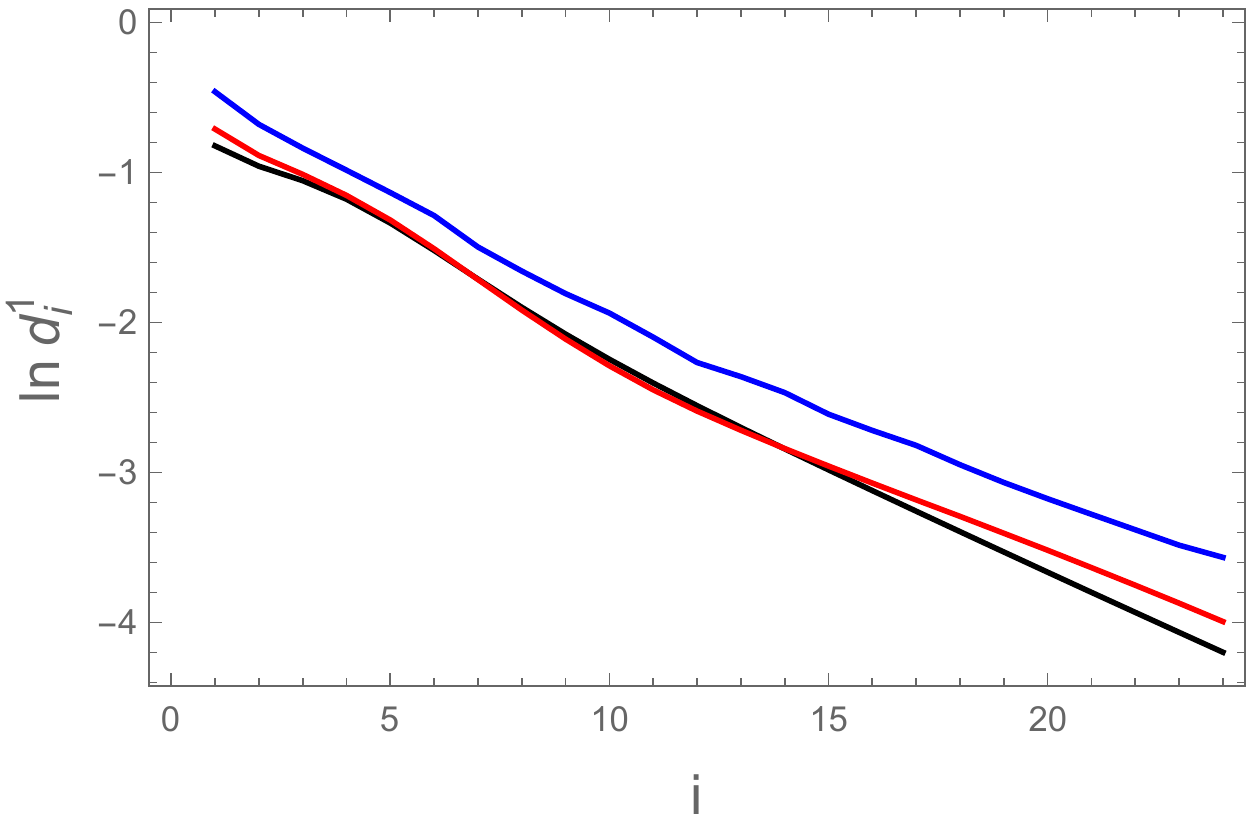,height=5.0cm}
\\
{\bf a} & {\bf b}
\end{tabular}
\caption{The shifted difference norms ({\bf a})  $\delta^{1}_i$ of state vectors  
and ({\bf b}) $d^{1}_i$ of cell state vectors of LSTM network with $n=10$
decay exponentially with $i$ for the noise amplitude
$a=0$ (black), $0.15$ (red) and $0.9$ (blue).}
\label{Fig3}
\end{center}
\end{figure}

In the LSTM network the relations similar to (\ref{pred0_i+1}-\ref{delta_i_gen}) are valid
with respect to the cell state vectors $\bm c^j_i$ and one can write for 
$\bm d_i^{j} = {\bm c}^{j}_{i+1}-{\bm c}^{j+1}_{i}$:
\be
\bm d_{m-1}^{j} = \bm B^j \cdot \bm d_0^{j},
\quad
\bm B^j = \prod_{k=0}^{m-1} \bm B^{j}_k,
\quad 
\bm B^{j}_k =  \frac{\partial \bm{\mathcal F}({\bm  X}^{j}_{k+1}, \bm c = {\bm c}^{j+1}_{k-1})}{\partial \bm c} ,
\quad 
1 \le j \le p.
\label{d_i_gen} 
\ee
The computations for $j=1$ show (see Figure \ref{Fig3}) that both 
$\delta^1_i$ and $d^1_i$  decrease exponentially with $i$
\be
\delta^{1}_i = \delta^{1}_1 e^{-\alpha i},
\quad\quad
d^{1}_i = d^{1}_1 e^{-\beta i},
\label{D_i}
\ee
and both
decay rates $\alpha$ and $\beta$  are not affected by the noise strength
but depend on $i$, i.e., for large $i$ they might tend to zero.
It is possible that decay rates behavior also depends on the 
number of neurons $n$.
The simulations show that similar behavior remains valid for all steps of the 
prediction procedure
\be
\delta^{j}_i \sim e^{-\alpha i},
\quad\quad
d^{j}_i \sim e^{-\beta i},
\quad
1 \le j \le p.
\label{D_i_gen}
\ee

This means also that the state vector ${\bm s}^{j+1}_{m-1}$ (next to last in the 
sequence $\bm S^{j+1}$) is very close to the last vector ${\bm s}^{j}_{m}$ 
of the preceding sequence $\bm S^{j}$, i.e., 
\be
{\bm s}^{j+1}_{m-1} = {\bm s}^{j}_{m} + \bm\epsilon^j,
\quad\quad
\epsilon^j \ll 1.
\label{last_state_small}
\ee

\subsection{Approximate governing transformation}

Now it is time to recall that the state vector ${\bm s}^{j}_{m}$
gives rise to the prediction $\bar{\bm x}_{m+j} = \bm W\cdot {\bm s}^{j}_{m}+\bm b$,
and this value is used as the last element of the input sequence for the next prediction step:
$\bm X^{j+1}_m = \bm W\cdot {\bm s}^{j}_{m}+\bm b$.

Employ the relation (\ref{map0}) for $i=m$ to find
\be
{\bm s}^{j+1}_{m} =  
\bm{\mathcal F}({\bm  X}^{j+1}_{m}, {\bm s}^{j+1}_{m-1}) = 
\bm{\mathcal F}( \bm W\cdot {\bm s}^{j}_{m}+\bm b, {\bm s}^{j}_{m} + \bm \epsilon^j) 
\approx 
\bm{\mathcal F}( \bm W\cdot {\bm s}^{j}_{m}+\bm b, {\bm s}^{j}_{m}) = 
\bm{\mathcal G}({\bm s}^{j}_{m}).
\label{mapnew}
\ee
The map $\bm{\mathcal G}$ for LSTM is defined by the transformations  (for $j > 1$)
\bea
& 
{\bm s}^{j}_{m}  =  {\bm o}^{j}_{m} \otimes \tanh {\bm c}^{j}_{m},
&
{\bm c}^{j}_{m}  =  {\bm f}^{j}_{m} \otimes {\bm c}^{j-1}_{m} + {\bm i}^{j}_{m} \otimes {\bm m}^{j}_{m},
\nonumber \\
&
{\bm o}^{j}_{m}  =  \sigma(\tilde{\bm W}_{os}{\bm s}^{j-1}_{m} + \tilde{\bm b}_o),
&
{\bm i}^{j}_{m}  =  \sigma(\tilde{\bm W}_{is}{\bm s}^{j-1}_{m} + \tilde{\bm b}_i ),
\label{LSTMapprox} \\
&
{\bm f}^{j}_{m}  =  \sigma(\tilde{\bm W}_{fs}{\bm s}^{j-1}_{m} + \tilde{\bm b}_f ),
&
{\bm m}^{j}_{m}  =  \tanh(\tilde{\bm W}_{ms}{\bm s}^{j-1}_{m} + \tilde{\bm b}_m),
\nonumber
\eea
where 
\be
\tilde{\bm W}_{as} = \bm W_{ax}\cdot\bm W+ \bm W_{as},
\quad
\tilde{\bm b}_{a} =  \bm W_{ax}\cdot\bm b + \bm b_{a},
\quad
a = i,f,m,o,
\label{LSTMnew}
\ee
and $\bm s^1_m$ and $\bm c^1_m$ are obtained by application of
(\ref{LSTM}) to the original input sequence.
It is easy to see that (\ref{LSTMapprox}) can be obtained 
from (\ref{LSTM}) by setting all $\bm W_{ax} = \bm 0$ and using the replacements
$\bm W_{as} \to \tilde{\bm W}_{as}$ and $\bm b_{a} \to \tilde{\bm b}_{a}$ defined in 
(\ref{LSTMnew}).
Similar procedure can be applied to (\ref{basic}) and (\ref{Fgated}) for basic and gated RNN respectively
and it gives for the  basic network a simple transformation
\be
{\bm s}^{j}_{m} =
\tanh(\tilde{\bm W}_{is}\cdot {\bm s}^{j-1}_{m}+\tilde{\bm b}_i).
\label{mapnewb} 
\ee
We observe that the influence of the input sequence $\bm X^j$ (and the noise
contained in it) on
the dynamics of the RNN last state vector $\bm s^j_m$ is negligible and the latter
is almost completely determined by the same vector $\bm s^{j-1}_m$ at the preceding prediction step. 


\section{A new fast algorithm for trajectory prediction}
\label{algorithm}

The main result in previous Section implies that after computation of $\bm s^1_m$ 
using $m$ times the recursion (\ref{map0}) the original input sequence can be dropped
and the transformation (\ref{mapnew}) is applied recursively $p-1$ times to generate
$\bm s^j_m$ for $2 \le j \le p$. Then the linear transformation 
(\ref{linear}) produces the desired sequence $\bar{\bm x}_{m+j}$ for $1 \le j \le p$.

\begin{figure}[h!]
\begin{center}
\psfig{figure=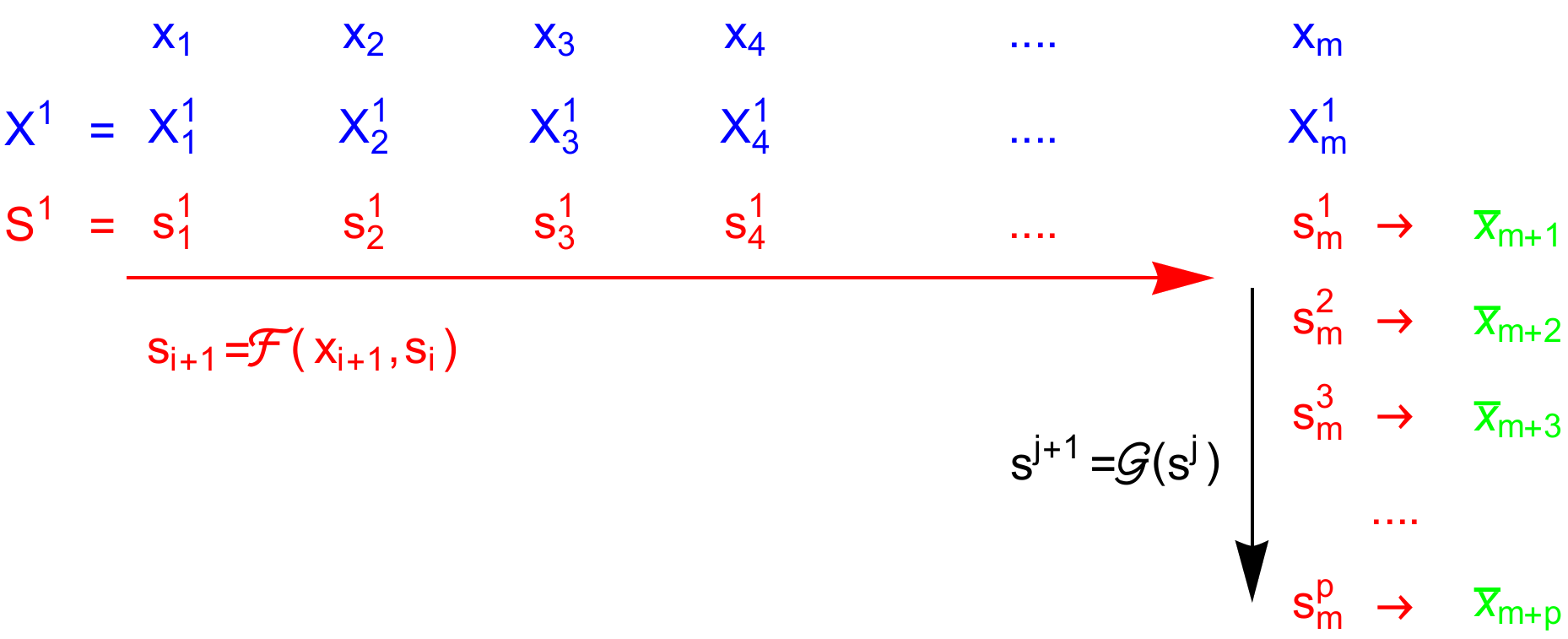,height=5.0cm} 
\caption{The approximate scheme of the recursive prediction based on (\ref{mapnew}).
The standard prediction sequence (\ref{map0}) is evoked only once to produce
$\bm s^1_m$ and then the approximate algorithm (\ref{mapnew}) is applied recursively
to produce $\bm s^j_m$ (red). The predicted points $\bar{\bm x}_{m+j}$ (green) are computed using
(\ref{linear}).}
\label{Fig5}
\end{center}
\end{figure}

These steps represent a new fast prediction algorithm (Figure \ref{Fig5}).
The transformation (\ref{mapnew}) for the untrained network with 
randomly chosen parameters might produce non-smooth and even chaotic dynamics
but even in this case the noise component in the input sequence plays no role
in the generation of the anticipated points. On the other hand this noise component
can strongly affect the result of RNN training influencing 
the weights and biases of the trained network.

\begin{figure}[h!]
\begin{center}
\begin{tabular}{cc}
\psfig{figure=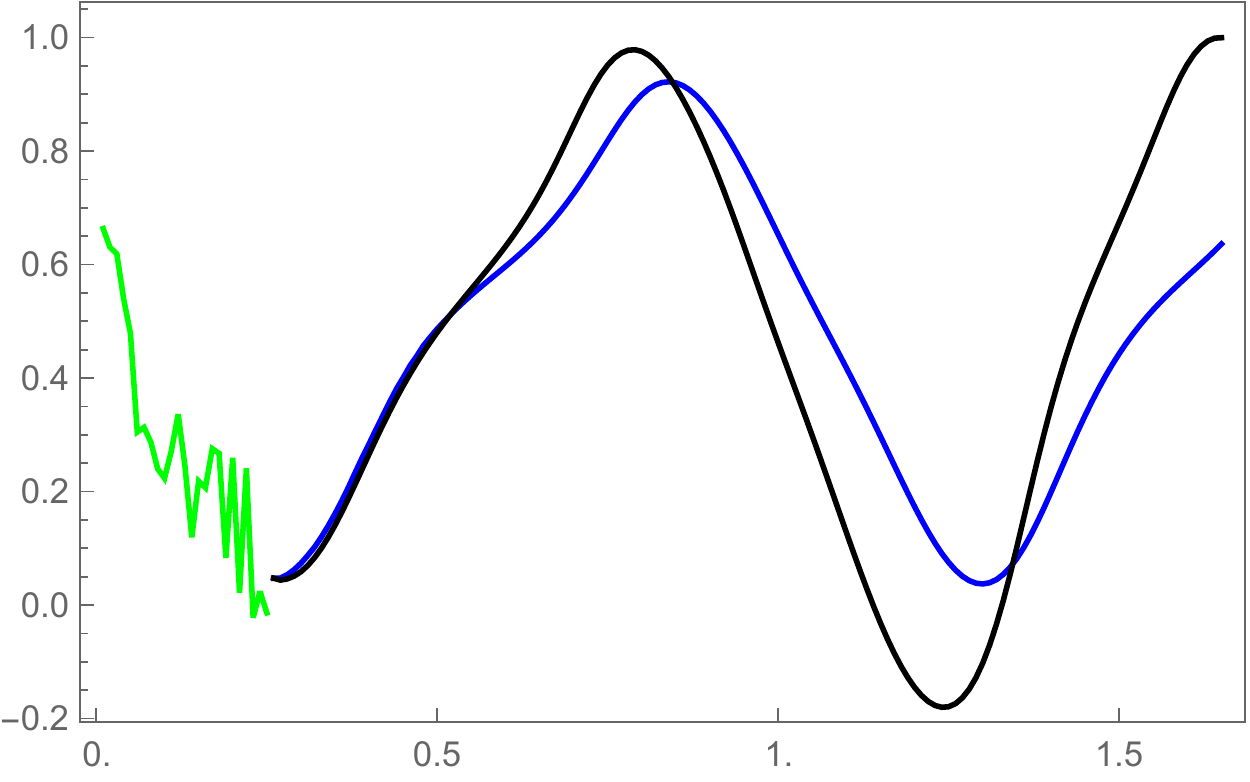,height=4.0cm}  &
\psfig{figure=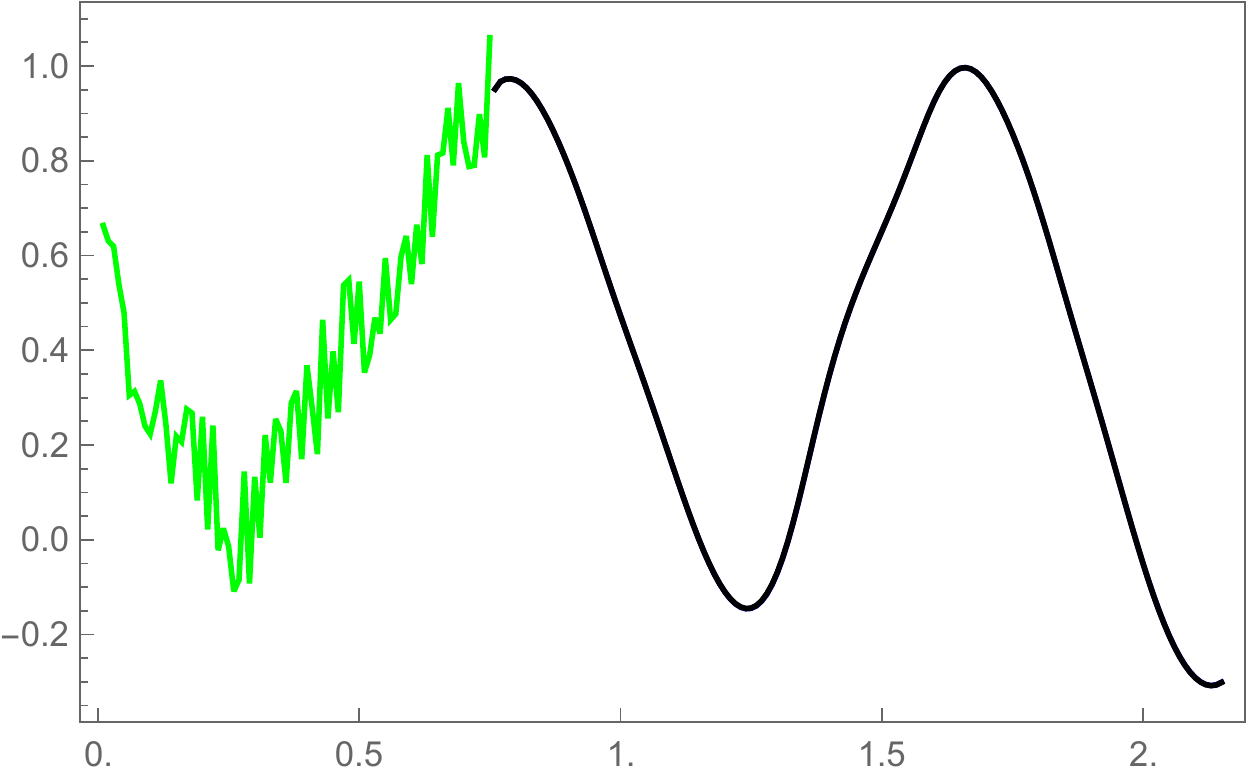,height=4.0cm}
\\
{\bf a} & {\bf b} \\
\psfig{figure=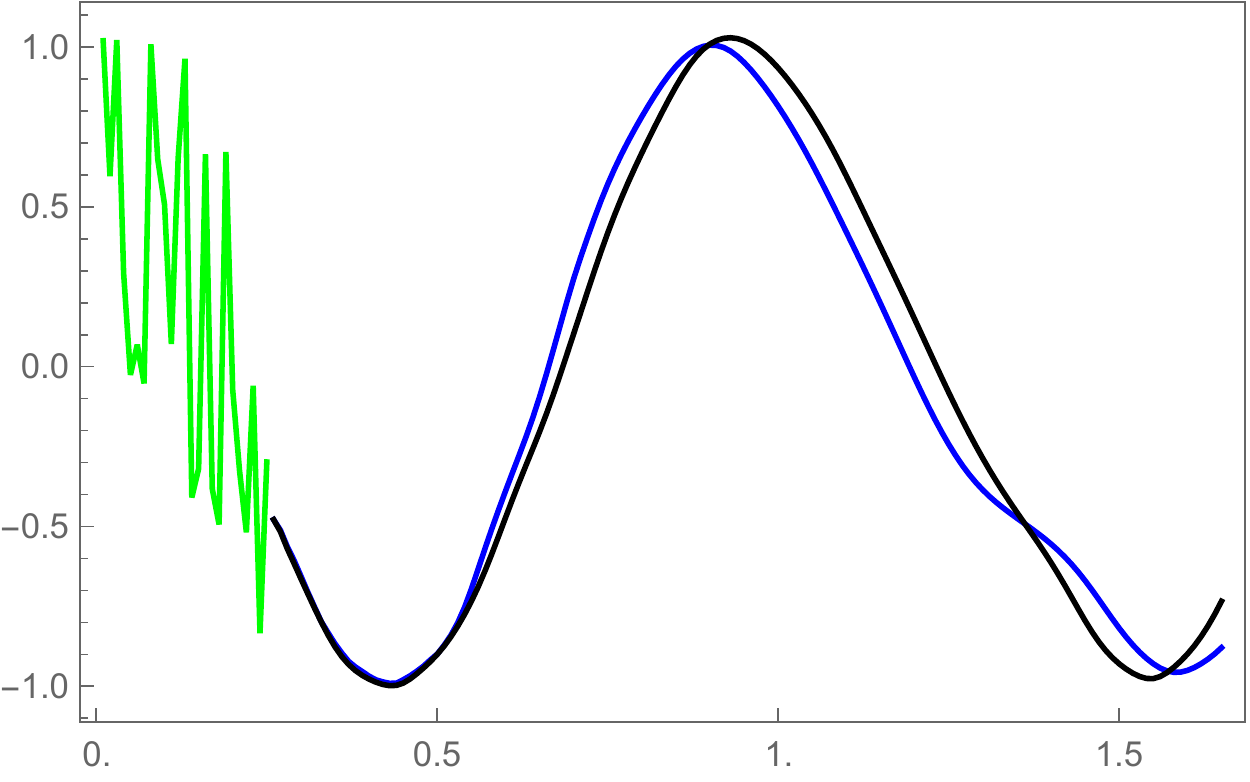,height=4.0cm}  &
\psfig{figure=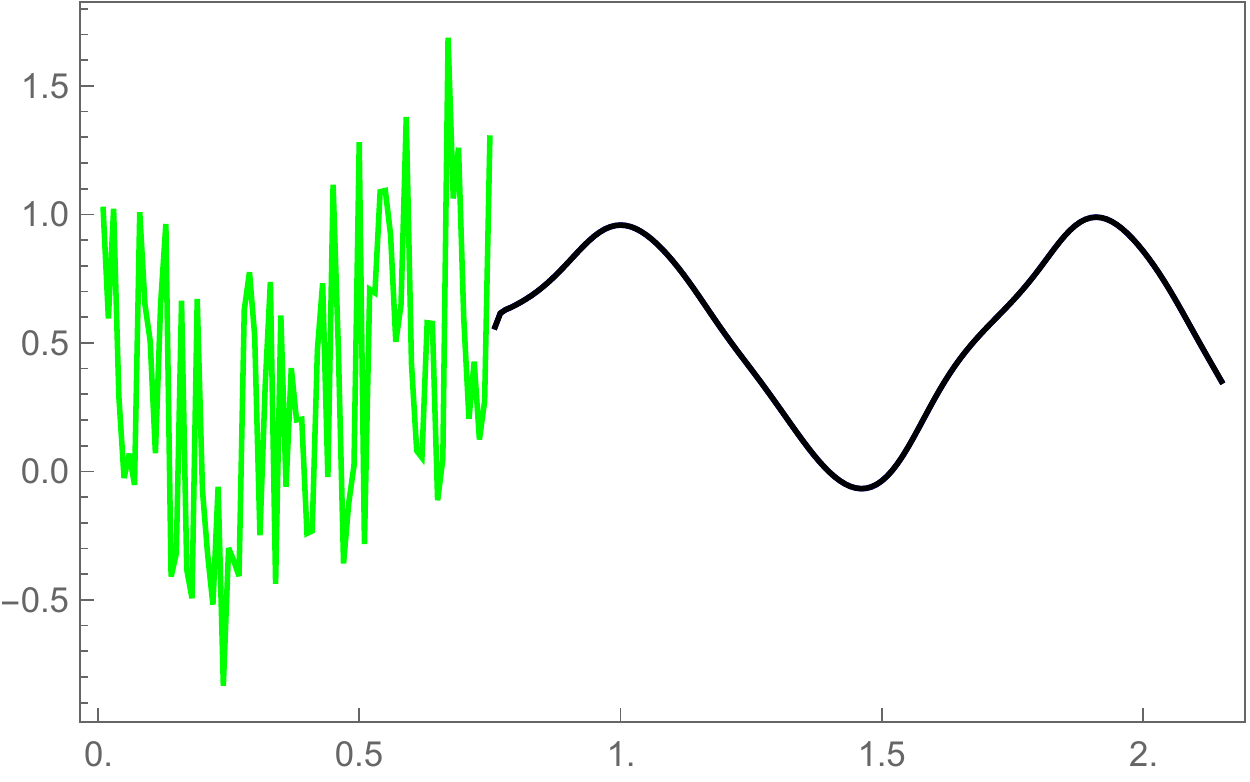,height=4.0cm}
\\
{\bf c} & {\bf d}
\end{tabular}
\caption{Comparison of the predictions for the
trained LSTM network by the moving window procedure (blue) and by using the 
map (\ref{LSTMapprox}) (black) for the triangle wave input
sequence (green) with variable noise amplitude $a$ and length $m$:
({\bf a}) $a=0.15,\ m=25$,  
({\bf b}) $a=0.15,\ m=75$,
({\bf c}) $a=0.9,\ m=25$,  
({\bf d}) $a=0.9,\ m=75$; in ({\bf b}) and ({\bf d})  both predictions coincide.
}
\label{Fig4}
\end{center}
\end{figure}

We use the approximate map (\ref{mapnew}) to compute the predicted sequence
for the input of different length $m$ and compare the results to the prediction
made by iterative application of RNN. We find that increase in input sequence length $m$ improves 
the approximate prediction (Figure \ref{Fig4}) up to a perfect coincidence with the 
traditional approach result. It is explained by the fact that for large $m$ the 
difference $\bm\epsilon^1$ becomes extremely small that increases the 
accuracy of the map (\ref{mapnew}).
Moreover, when we increase the input sequence noise amplitude $a$ six times 
compared to the value at which LSTM network was trained, the approximate procedure still 
generates prediction coinciding with the one produced by LSTM itself (Figure \ref{Fig4}d).

\begin{figure}[h!]
\begin{center}
\begin{tabular}{cc}
\psfig{figure=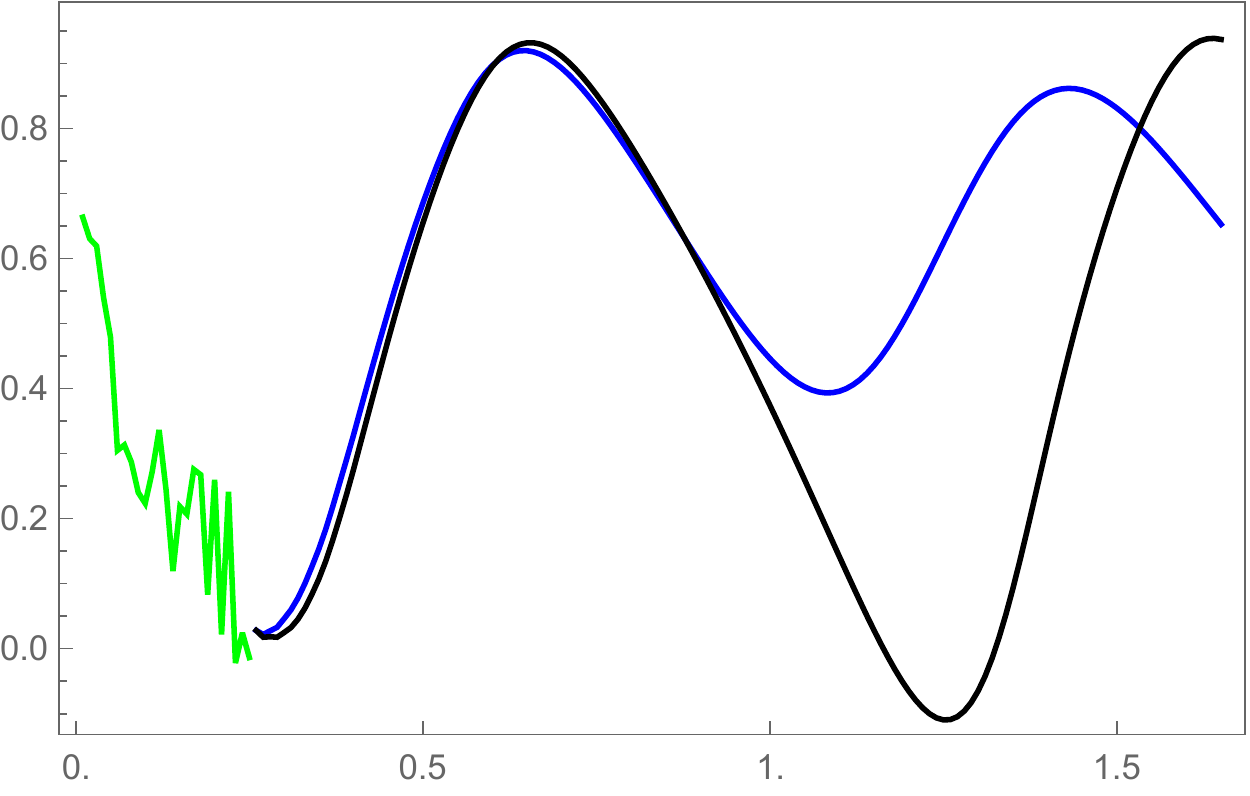,height=4.0cm}  &
\psfig{figure=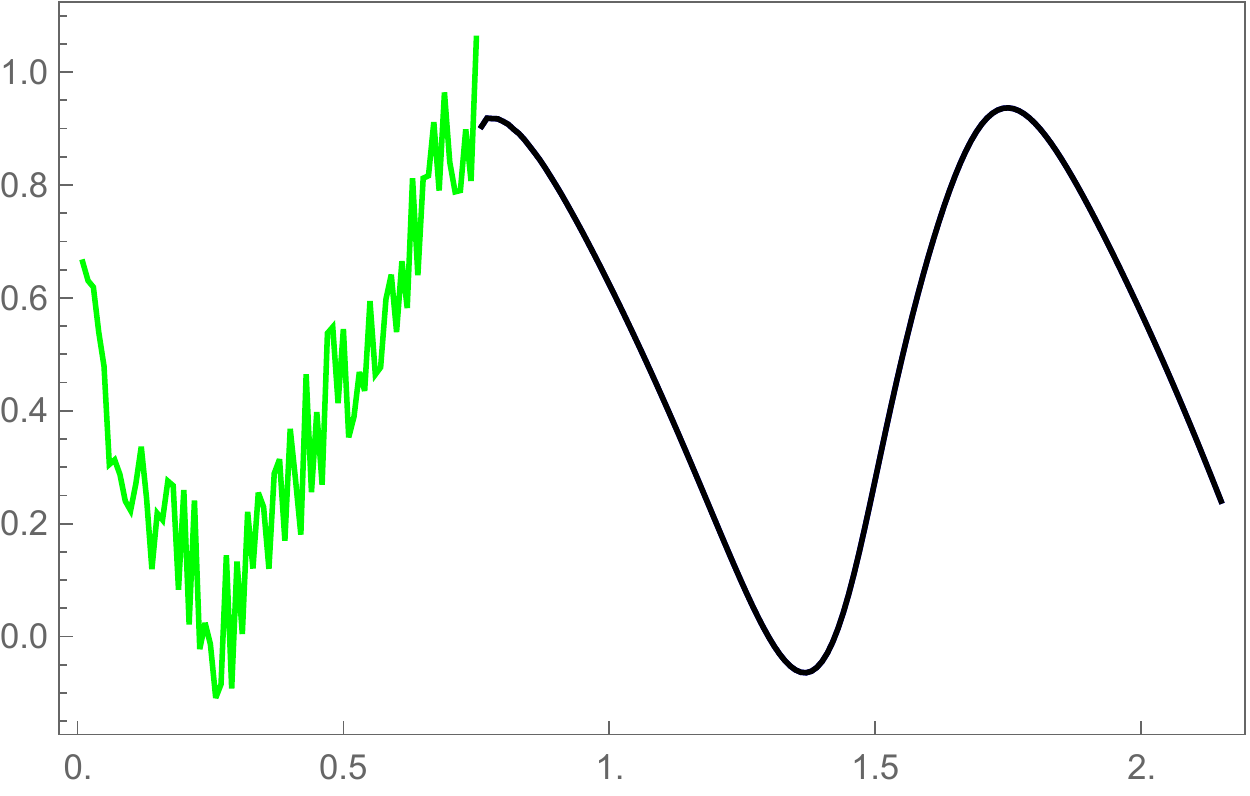,height=4.0cm}
\\
{\bf a} & {\bf b}
\end{tabular}
\caption{Comparison of the predictions for the basic
RNN with $n=10$  by the moving window algorithm (blue) and by the map (\ref{mapnewb}) (black) for the triangle wave input
sequence (green) of  ({\bf a}) $m=25$ and ({\bf b}) $m=75$ points with noise amplitude $a=0.15$;
in ({\bf b})  both predictions coincide.}
\label{FigB2}
\end{center}
\end{figure}

We also compare the predictions made by the RNN governed by
(\ref{basic}) and (\ref{mapnewb}) and find that these predictions
coincide for large $m$ (Figure \ref{FigB2}b).

We observe that the moving window prediction generating $p$ time series points using the trained RNN is
a recursion ($p$ times) each consisting of $m$ inner recursions, i.e., total $R_o=mp$ recursion steps
while the approximate procedure (\ref{mapnew}) 
replaces it by $R_a = m+p-1$ recursions (Figure \ref{Fig5}). 
Assuming that the computation time ${\cal T}$ is linearly proportional to the total recursion number
${\cal T} = \mu R$ estimate a speed up $\kappa = {\cal T}_o/{\cal T}_a$.
The length $m$ of the input sequence $\bm X$ should be quite large ($m \gg 1$) in order to generate
a high quality prediction. The length $p=\gamma m$ of the predicted sequence  $\bar{\bm X}$ is comparable to
$m$, i.e., $\gamma \gtrapprox 1$ and we find the estimate of the prediction times ratio 
$\kappa = mp/(m+p)  = \gamma m/(1+\gamma) > m/2$. Thus the approximate prediction algorithm gain
is proportional to the length of the input sequence. 
We observed that $m \approx 50$ leads to high quality of the approximate scheme (Figures \ref{Fig4}, \ref{FigB2}) and 
thus one can have speed up of an order of magnitude without loss of prediction quality.

\section{Algorithms robustness analysis}
\label{neuro}

The results presented above can have important implications in neuroscience.
If one assumes that brain uses recurrent networks for trajectory prediction and it employs 
the moving window procedure described in Section \ref{algorithm0} (see Figure \ref{Fig0}) then 
the implementation of this algorithm requires satisfaction of several conditions.
These include -- 
the value (amplitude) of the input element should not change significantly during 
time interval when this element is used for prediction; the order of the elements of the input sequence at the 
second and subsequence steps of prediction should not change. The first condition
can be broken if the signal value is perturbed by inner noise or it decays.
As the influence of noise on the input sequence is shown not to be critical for the prediction
we will focus on the signal decay influence of the prediction quality.
The second condition is probably more difficult to satisfy and we have to consider a case when 
on each step of prediction some elements of the updated input sequence are 
partially reshuffled.

Consider first how the input element decay rate affects the quality of prediction.
\begin{figure}[h!]
\begin{center}
\begin{tabular}{cc}
\psfig{figure=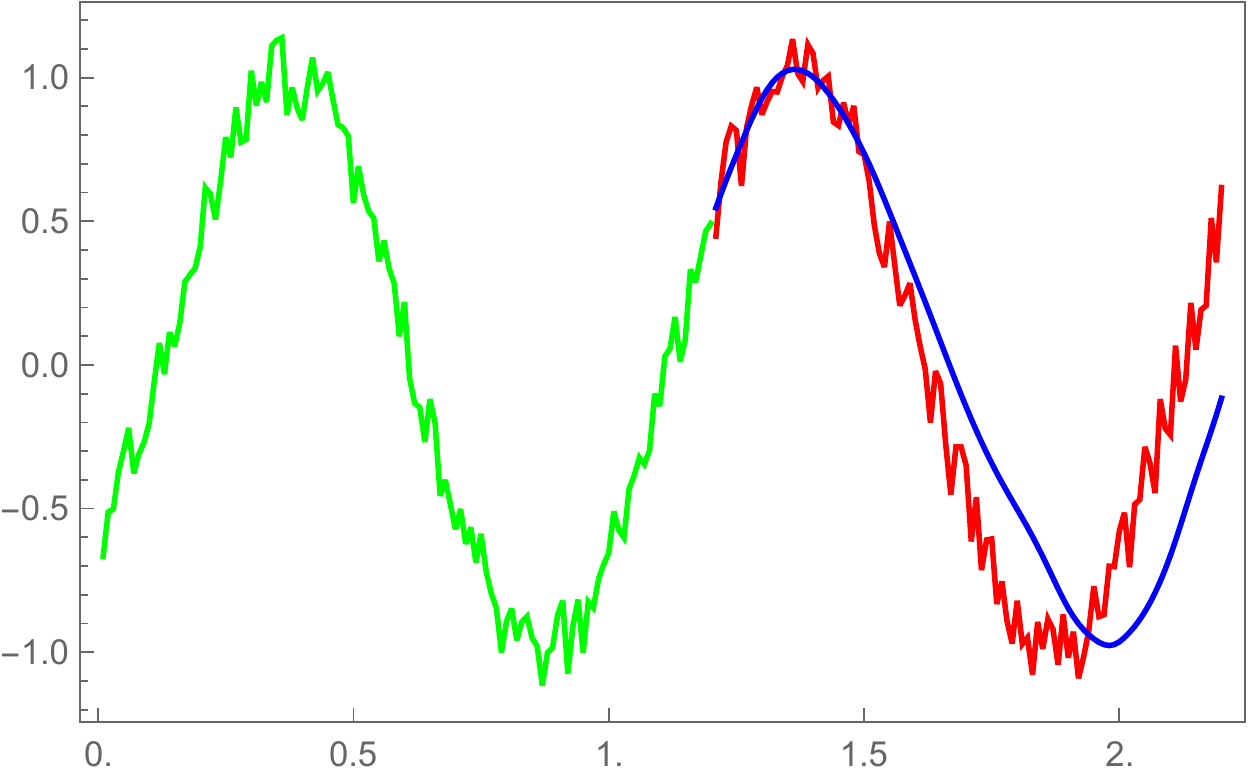,height=4.5cm}  &
\psfig{figure=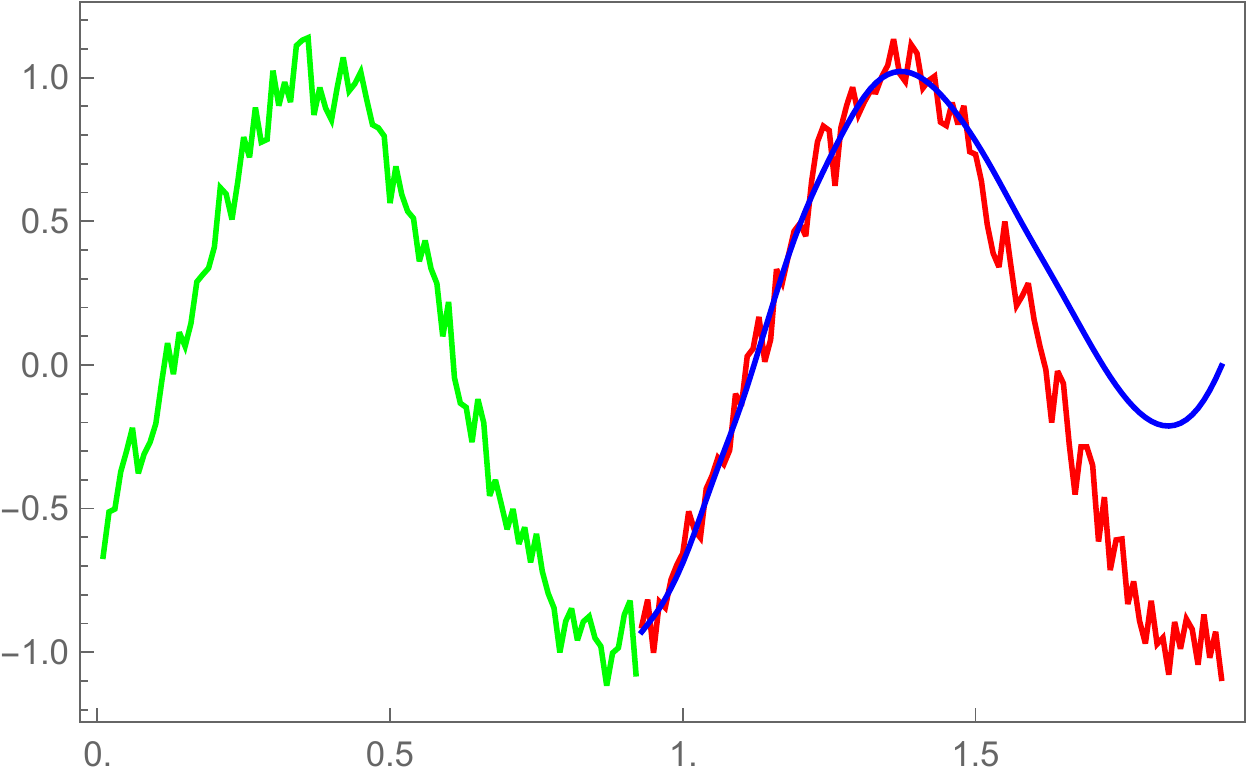,height=4.5cm}
\\
{\bf a} & {\bf b}\\
\psfig{figure=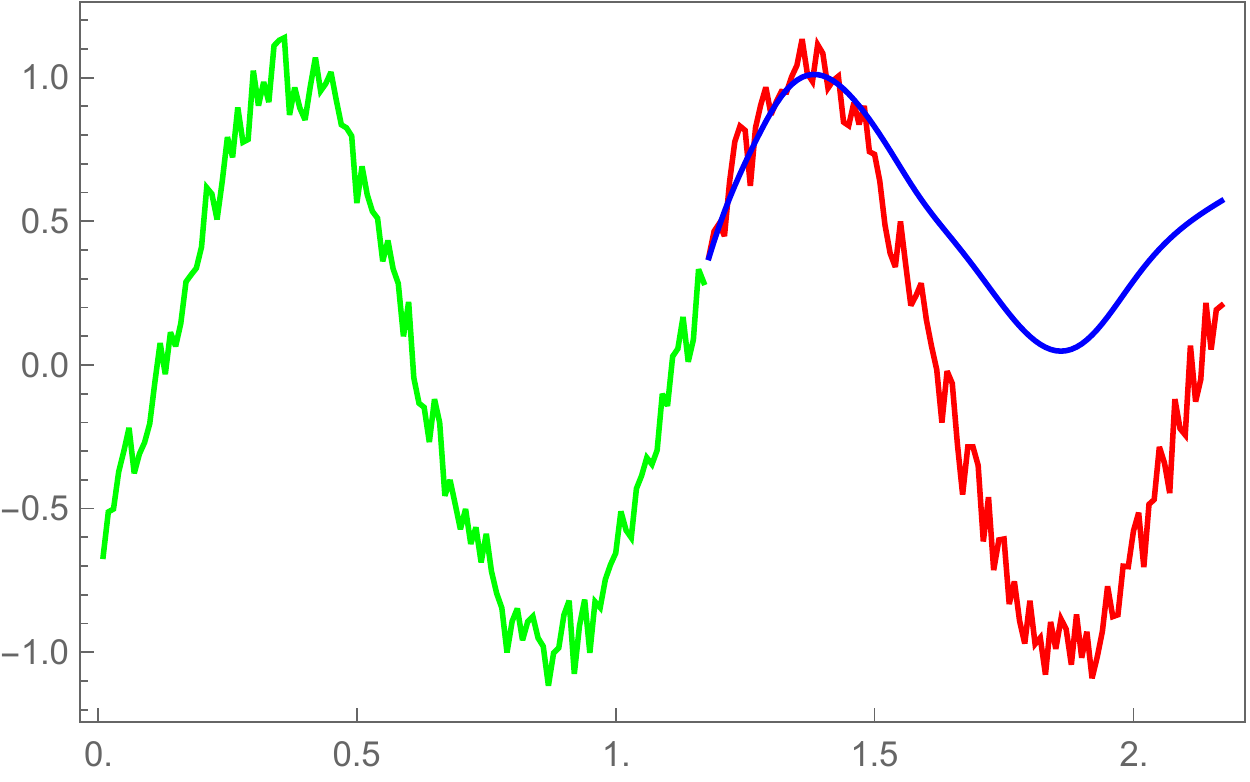,height=4.5cm}  &
\psfig{figure=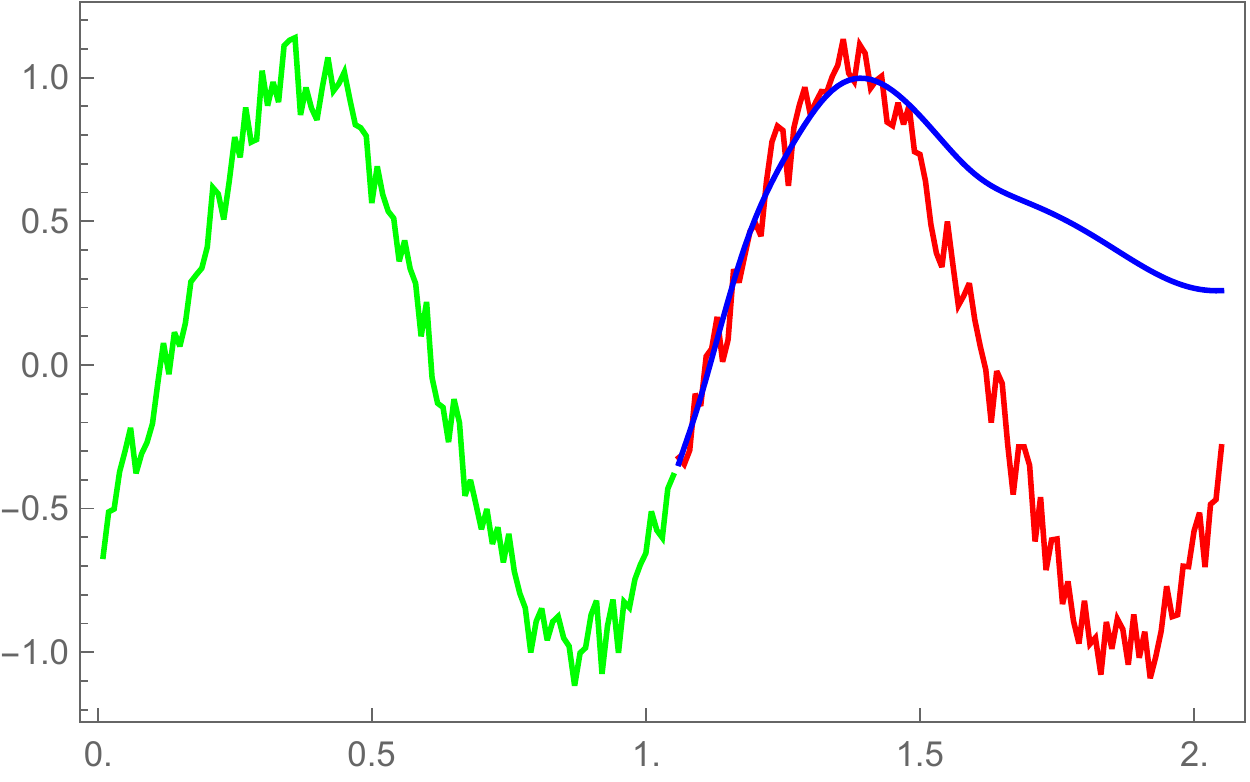,height=4.5cm}
\\
{\bf c} & {\bf d}
\end{tabular}
\caption{Comparison of the predictions by the LSTM
network (blue) with $n=10$ to the continuation (red) of the sine wave input (green)
for different rate $\alpha$ of input values exponential decay: 
({\bf a}) $0$, ({\bf b}) $0.002$, ({\bf c}) $0.005$, ({\bf d}) $0.008$.}
\label{FigBrain1}
\end{center}
\end{figure}
For the LSTM network we observe that the increase of the decay rate leads to faster deviation of the predicted 
trajectory from the expected one (Figure \ref{FigBrain1}), nevertheless the predicted trajectory remains quite 
smooth.

\begin{figure}[h!]
\begin{center}
\begin{tabular}{cc}
\psfig{figure=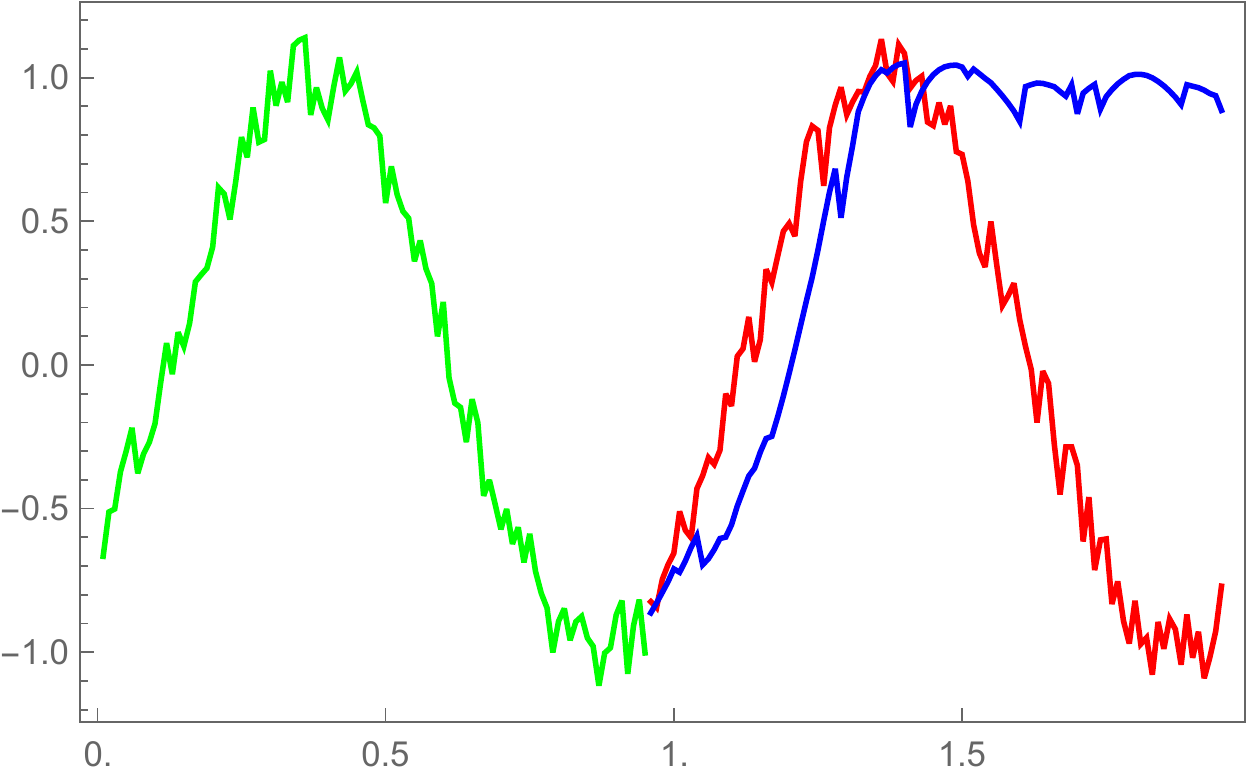,height=4.5cm}  &
\psfig{figure=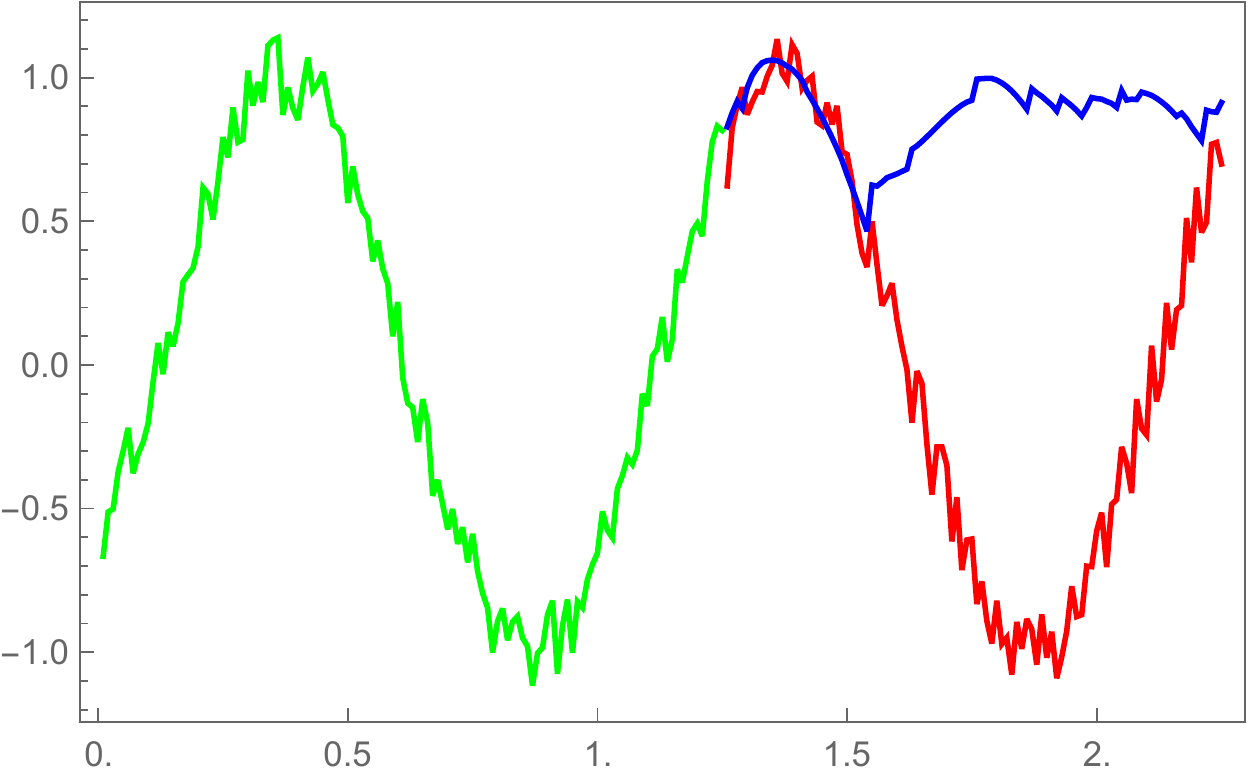,height=4.5cm}
\\
{\bf a} & {\bf b}\\
\psfig{figure=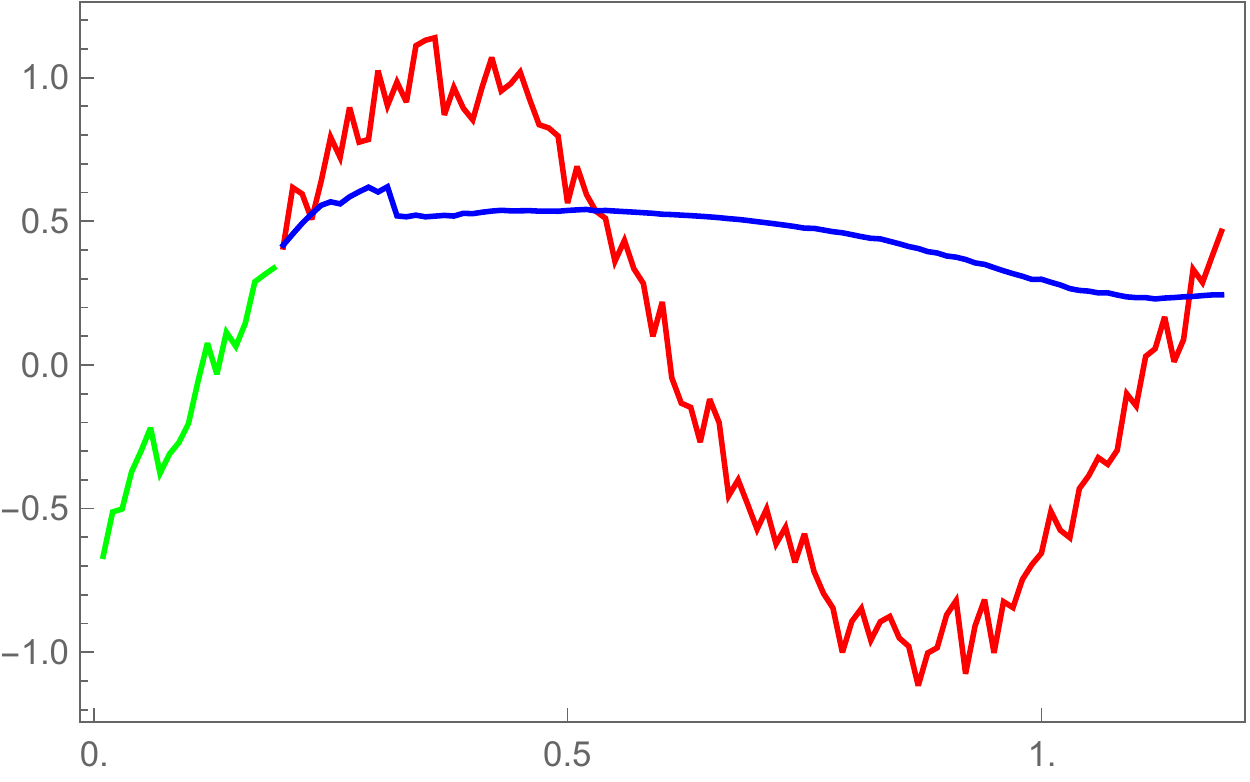,height=4.5cm}  &
\psfig{figure=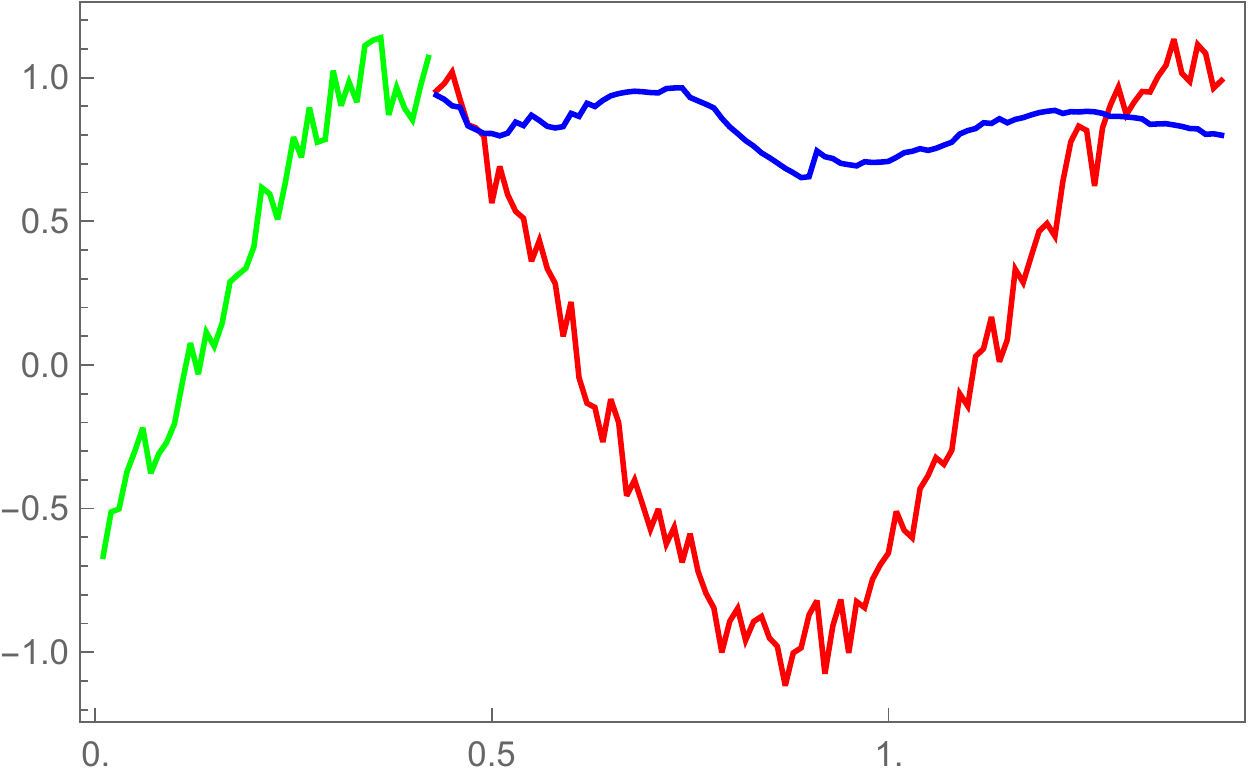,height=4.5cm}
\\
{\bf c} & {\bf d}
\end{tabular}
\caption{Comparison of the predictions by the trained LSTM
network (blue) with $n=10$ to the continuation (red) of the sine wave input (green)
when the input sequence is partially reshuffled.}
\label{FigBrain2}
\end{center}
\end{figure}

It appears that the partial reshuffling of the input sequence at each prediction step affects not only
the prediction quality but also generates nonsmooth extrapolated curves (Figure \ref{FigBrain2}).
We observe that satisfaction of both conditions mentioned above is critical for 
a successful prediction using the moving window algorithm and if any of them is not satisfied 
the increase of the length $m$ of input sequence makes an accurate prediction nearly impossible.
The reduced algorithm 
(see Figure \ref{Fig5}) is free of these limitations as it does not require 
any knowledge of 
the input sequence $\bm X^j$ for $j > 1$ but instead employs the internal
network dynamics, and the quality of prediction grows with the length $m$ of the initial input sequence.
Thus we conclude that the new fast algorithm appears to be much more robust
compared to the traditional moving window approach.


\section{Discussion}
\label{discussion}

In this manuscript we show that the predictive RNN based on a single 
recurrent layer with a small number of neurons works as an effective noise filter.
Namely, when the RNN is supplied by the noisy input sequence of 
(multidimensional) time series points and used recursively for series 
extrapolation it generates points that belong to some smooth curve
that mimics the smoothed original time series. 
Using the analysis of the recursive prediction procedure 
we established a set of
conditions required to observe such behavior. 
These conditions imply that the governing transformation
of the predictive algorithm reduces to one that requires the 
input sequence only once and later 
does not depend on it. As the result the
predictive algorithm can be drastically simplified and accelerated 
without loss of accuracy. The overall quality of prediction strongly depends
on the length of  the input sequence while the acceleration is proportional 
to it. Thus using the approximate predictive algorithm
one can both increase the quality and save time and computational resources.

These results allow to conclude that RNNs with 
several recurrent layers of a single or multiple types would have the same 
property of noise filtration off an input sequence. 
Moreover it is possible to suggests that any neural network of
several layers would share this behavior if it has a recurrent network preceding
a last layer that generates the network prediction.

The approximate predictive algorithm is governed by a multidimensional  discrete map
with the parameters determined by the weights and biases of the trained RNN only and does not
require the input sequence.
In all our numerical experiments we observe that the parameters of the trained 
network always lead to smooth dynamics generated by this reduced map. The same time 
setting these parameters to random real values sometimes produces nonsmooth and 
quite nontrivial dynamics including complex periodic and even chaotic trajectories.
It is very important to understand what is special 
about the parameters of the trained network that they {\it always} produce smooth 
trajectory generated by both the original and approximate predictive schemes.

Another important aspect of RNN noise filtering is related to neuroscience.
Brain ability to predict a trajectory is 
one of the most important requirements for survival and this 
natural ability is highly developed. By default the brain should be able to predict 
trajectories based on incomplete or noisy data, and it has to do this with high reliability.
Moreover, the predictions should be made for several objects simultaneously and
it requires large resources.
Even if an object actual trajectory in space is smooth it is transferred into brain by 
the receptors as a {\it noisy} time series.
The trajectory prediction is usually considered as a two-stage process
-- first, the brain performs initial classification of the trajectory and then, in case
when the organism should somehow react to this specific motion,
a precise predictive mechanism is activated. If the available data is noisy
both these stages would require more resources compared to processing of 
smooth trajectories.
We hypothesize that first of all activates an additional inexpensive (with small number of neurons) 
recurrent network. It would effectively filter noise out and transfer a cleaned
smooth trajectory segment to the classification and then to precise predictive 
networks. Note that in this case the latter networks resources can be greatly 
reduced.

We also learned that the prediction process itself can be significantly accelerated 
by using the approximate algorithm described in the manuscript. 
It would be interesting to address 
a possibility of a physiological implementation of this scheme.
If this algorithm does work in the brain
the trajectory prediction is done in two stages -- first the 
existing trajectory segment is fed into the network and the first point is predicted.
Then the input information is forgotten and the brain predicts 
subsequent points based on the approximate scheme. 
We showed that the moving window prediction procedure is very sensitive to 
various perturbations of the input sequence during its update that might 
strongly reduce the prediction quality. Moreover, the longer is the input sequence
the higher chance is for these perturbations to influence the result.  
On the other hand, 
the reduced fast algorithm is much more robust with respect to those 
perturbations and allows to reach high predictability which is
proportional to the length of the  input sequence.

One has to take into account that the number of predicted elements
is usually smaller or approximately equal to the length of the input sequence
as the prediction accuracy is inversely proportional to the length of predicted sequence. 
Thus the receptors provide a new input sequence is 
and a correction of predicted trajectory is performed. 
It  saves resources and helps to resolve the problem of prediction time minimization  -- 
there exists a range of lengths $m$ of the input sequence for which 
the prediction quality is proportional to $m$ thus brain tends to 
increase the value of $m$. This increase requires a linearly proportional increase
in prediction time when the moving window algorithm is employed.
A switch to the approximate algorithm allows significant reduction
in the processing time without loss in the prediction quality.

\section*{Acknowledgements}
The author wishes to thank Vladimir Zverev, Yuri Shvachko, Mitya Chkolvsky and Kausik Si for fruitful discussions.

\newpage
\noindent
{\Large \bf Appendix}

\appendix

 
\section*{Shifted difference dynamics for basic recurrent network}

\setcounter{equation}{0}
\renewcommand{\theequation}{A\arabic{equation}}

The simplest RNN transformation reads
\be
\bm s_{i} = f (\bm W_{ix} \cdot \bm x_{i} + \bm W_{is} \cdot \bm s_{i-1} + \bm b_i),
\label{Fbasic}
\ee
where the nonlinear scalar function $f(x) = \tanh x$ is applied to all components of its vectorial argument.
In (\ref{Fbasic}) the index $i$ in $\bm W_{ix}, \ \bm W_{is}$ and  $\bm b_i$ stands for the input gate. 
The shifted difference $\bm \delta_{i}^j = \bm s_{i+1}^j - \bm s_{i}^{j+1}$  reads
\bea
\bm \delta_{i}^j &=&
 f(\bm W_{is} \cdot \bm s_{i}^j + \bm W_{ix} \cdot \bm x_{i+1}^j + \bm b)
-  f( \bm W_{is} \cdot \bm s_{i-1}^{j+1} + \bm W_{ix} \cdot \bm x_{i}^{j+1} + \bm b)
\nonumber \\
&=&
 f(\bm W_{is} \cdot \bm s_{i}^j + \bm W_{ix} \cdot \bm x_{i+1}^j + \bm b)
- f(\bm W_{is} \cdot (\bm s_{i}^{j} - \bm \delta_{i-1}^{j})+ \bm W_{ix} \cdot \bm x_{i+1}^{j} + \bm b)
\nonumber \\
&=&
f(\bm y_{i}^j) -  f(\bm y_{i}^j  - \bm W_{is} \cdot \bm \delta_{i-1}^{j})
\quad \quad 
\bm y_{i}^j = \bm W_{is} \cdot \bm s_{i}^j + \bm W_{ix} \cdot \bm x_{i+1}^j + \bm b = f^{-1}(\bm s_{i+1}^j),
\label{sdiff_basic0}
\eea
where $f^{-1}$ denotes an inverse function to $f$ and 
we use the relation $\bm x_{i+1}^j = \bm x_{i}^{j+1}$.
Assume that $|\bm W_{is} \cdot \bm \delta_{i-1}^{j}| \ll y_{i}^j$
and find in the lowest expansion order
$$
\bm \delta_{i}^j \approx
\frac{\partial f(\bm y_{i}^j)}{\partial \bm y_{i}^j} \cdot (\bm W_{is} \cdot \bm \delta_{i-1}^{j}),
$$
and $\bm M_{i}^j = \partial f(\bm y_{i}^j)/\partial \bm y_{i}^j$ is a square matrix
having the same dimensions as the matrix $\bm W_{is}$.
Recalling that the nonlinear transformation $f$ is actually a scalar function applied to 
all elements of its vector argument $\bm y_{i}^j$ one can write
for the  diagonal matrix $\bm M_{i}^j = \bm D[f'(\bm y_{i}^j)] \equiv
\mbox{diag} \{f'(\bm y_{i}^j)\}$.
This leads to
\be
\bm \delta_{i}^j \approx 
\bm A_{i}^j \cdot \bm \delta_{i-1}^{j},
\quad
\bm A_{i}^j = 
\bm D[f'(\bm y_{i}^j)] 
\cdot \bm W_{is} = 
\bm D[f'(f^{-1}(\bm s_{i+1}^j))] 
\cdot \bm W_{is}.
\label{sdiff_basic1}
\ee
For basic RNN $f'(x) = \tanh' x = \mbox{sech}^2 x,$ 
leading to $f'(f^{-1}(x)) = 1-x^2,$ and thus 
\be
\bm A_{i}^j = \bm D[1-\bm s_{i+1}^j\otimes\bm s_{i+1}^j] 
\cdot \bm W_{is}.
\label{sdiff_basic2}
\ee
Consider two vectors $\bm a,\ \bm b$ satisfying $\bm b = \bm A_i^j\cdot \bm a$.
Introduce a matrix $\bm U = (\bm A_i^j)^T\cdot \bm A_i^j$ where 
${}^T$ denotes matrix transposition and find its eigenvectors $\bm e_k$ 
and eigenvalues $\lambda_k$ satisfying
$\bm U\cdot\bm e_k = \lambda_k \bm e_k$.
Expand both vectors using the orthonormal basic $\{\bm e_k\}$ 
\be
\bm a = \sum_k w_{ak} \bm e_k, 
\quad
\bm b = \sum_k w_{bk} \bm e_k, 
\label{Umatr0}
\ee
and compute a squared norm of these vectors
\be
a^2 = \bm a\cdot\bm a = 
\sum_k w_{ak}^2,
\quad
b^2 = \bm b\cdot\bm b = 
\sum_k w_{bk}^2.
\label{Umatr1}
\ee
On the other hand we have
$$
\bm b\cdot\bm b = 
\bm a\cdot (\bm A_i^j)^T\cdot \bm A_i^j \cdot \bm a = 
\bm a\cdot \bm U \cdot \bm a,
$$
leading to 
\be
b^2 = 
\sum_k \lambda_k  w_{ak}^2.
\label{Umatr2}
\ee
Now choose $\bm a = \bm \delta_{i-1}^j$, $\bm b = \bm \delta_{i}^j$
and find the weights $w_{ak}$ in (\ref{Umatr0}).
The computations show that with these weights for majority of the recursion steps the relation 
(\ref{Umatr2}) gives $b^2 < a^2$,  i.e.,  $\delta_{i}^j <\delta_{i-1}^j$ 
so that for the trained 
RNN the transformation (\ref{sdiff_basic1}) is a contracting one which
implies the exponential decay of the shifted difference norm $\delta_{i}^j$.

\end{document}